\documentclass[journal]{IEEEtran}
\usepackage[american]{babel}
\usepackage{graphicx}
\usepackage{subfigure}
\usepackage{cite}
\usepackage{indentfirst}
\usepackage{multirow}
\usepackage{array}
\usepackage{amsmath}
\usepackage[numbers]{natbib}
\usepackage{color}

\newcommand{\sgn}{\operatorname{sgn}}

\ifCLASSINFOpdf

\else

\fi

\begin{document}
%
\title{ECML: An Ensemble Cascade Metric Learning Mechanism towards Face Verification}

%

\author{Fu~Xiong, Yang~Xiao, Zhiguo~Cao, Yancheng~Wang, Joey Tianyi~Zhou and Jianxin~Wu~\IEEEmembership{Member,~IEEE}~\thanks{Fu~Xiong, Yang~Xiao, Zhiguo~Cao, Yancheng~Wang are with National Key Laboratory of Science and Technology on Multispectral Information Processing, School of Artificial Intelligence and Automation, Huazhong University of Science and Technology, Wuhan 430074, China. E-mail: {xiongfu, Yang$\_$Xiao, zgcao, yancheng$\_$wang}@hust.edu.cn.}
\thanks{Joey Tianyi~Zhou is with Institute of High Performance Computing, A*STAR, Singapore. E-mail: {zhouty@ihpc.a-star.edu.sg.}}
\thanks{Jianxin~Wu is with the National Key Laboratory for Novel Software Technology, Nanjing University, Nanjing, 210023, P.R. China. E-mail: {wujx@lamda.nju.edu.cn.}}
\thanks{Yang Xiao is the corresponding author of this paper.}
\thanks{Manuscript received April 19, 2005; revised August 26, 2015.}}

\markboth{Journal of \LaTeX\ Class Files,~Vol.~14, No.~8, August~2015}%
{Shell \MakeLowercase{\textit{et al.}}: Bare Demo of IEEEtran.cls for IEEE Journals}

\maketitle
\begin{abstract}
Face verification can be regarded as a 2-class fine-grained visual recognition problem. Enhancing the feature's discriminative power is one of the key problems to improve its performance. Metric learning technology is often applied to address this need, while achieving a good tradeoff between underfitting and overfitting plays the vital role in metric learning. Hence, we propose a novel ensemble cascade metric learning (ECML) mechanism. In particular, hierarchical metric learning is executed in the cascade way to alleviate underfitting. Meanwhile, at each learning level, the features are split into non-overlapping groups. Then, metric learning is executed among the feature groups in the ensemble manner to resist overfitting. Considering the feature distribution characteristics of faces, a robust Mahalanobis metric learning method (RMML) with closed-form solution is additionally proposed. It can avoid the computation failure issue on inverse matrix faced by some well-known metric learning approaches (e.g., KISSME). Embedding RMML into the proposed ECML mechanism, our metric learning paradigm (EC-RMML) can run in the one-pass learning manner. Experimental results demonstrate that EC-RMML is superior to state-of-the-art metric learning methods for face verification. And, the proposed ensemble cascade metric learning mechanism is also applicable to other metric learning approaches.
\end{abstract}


\IEEEpeerreviewmaketitle

\section{Introduction}

\IEEEPARstart{F}{ace} verification is a long-term but still challenging research topic within computer vision and image processing community~\cite{Bianco2017faceverification,Sengupta2016faceverification,du2014discriminative}. It is of wide-range applications, such as public security system, human-machine interaction, e-commercial trading, etc~\cite{osadchy2010scifi,Ren2013A}. From pattern recognition perspective, face verification can be regarded as a 2-class fine-grained visual pattern recognition task to decide whether 2 face images indicate the same person~\cite{FVface}. It suffers from the challenges of high intra-person variation on illumination, pose, expression, age and occlusion. The central idea for improving the performance is to reduce intra-person variation, while enlarging inter-person difference. Metric learning on visual feature (e.g., CNN feature~\cite{Bhattarai2016CP}) is one of the commonly used technologies to address this.

Metric learning aims to learn a discriminative distance function towards the specific task. The yielded distance metric is able to enlarge inter-class distance, and reduces intra-class distance simultaneously. Besides face verification, metric learning is also widely applied to other fine-grained visual recognition tasks (e.g., person re-identification~\cite{dikmen2010pedestrian}, plant trait characterization~\cite{lu2017towards}, and {object detection~\cite{you2014local,du2016beyond}}).

 \begin{figure}
    \centering
    \includegraphics[width=8.8cm]{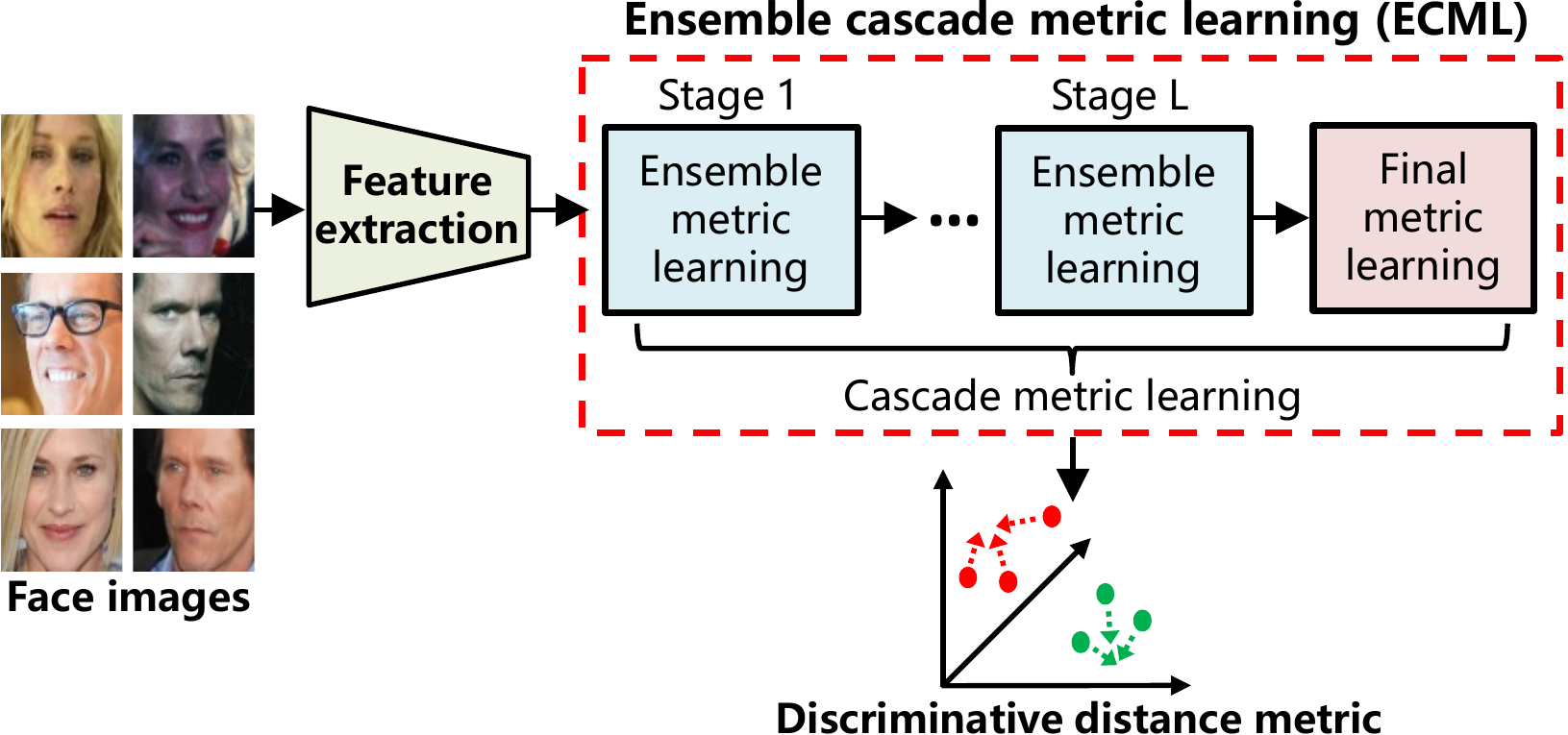}
    \caption{The main technical pipeline of the proposed ensemble cascade metric learning (ECML) mechanism towards face verification.}
     \label{fig:pipeline}
 \end{figure}

During the past decades, numerous efforts~\cite{KISSme,XQDA,xiong2014kernelmetric} have been paid on the development of metric learning technology from the different theoretical perspectives. Generally, they can be categorized into 2 families. One is linear paradigm ~\cite{KISSme,XQDA} that learns a linear distance metric transformation matrix $M$. Then, the distance between 2 samples $x_{i}$ and $x_{j}$ is calculated as $d_{ij}=(x_i-x_j)^TM(x_i-x_j)$. On the other hand, nonlinear model (e.g., {kernel-based manner~\cite{xiong2014kernelmetric,wang2019incorporating}}) is also studied to capture the nonlinearity of data. All of the linear and nonlinear approaches above are of shallow learning manner. Essentially, this leads to the fact that they may suffer from the underfitting problem. {To search more reliable metric, some works resort to the ensemble of multiple shallow metrics~\cite{dong2017dimensionality,paisitkriangkrai2015learning}. However, the plain ensemble manner cannot effectively alleviate the intrinsic underfitting problem within the shallow metrics. That is, the yielded metric may not be discriminative enough to reveal subtle difference among categories}.

Recently, the ideas of deep metric learning~\cite{Yi2014Deep,Hoffer2014Deep} have emerged. Leveraging the strong fitting capacity of deep neural network (DNN) ~\cite{VGG}, these deep models significantly outperform the shallow learning counterparts. In spite of the remarkable performance enhancement, large number of training samples are required to avoid overfitting ~\cite{lowshot}. Nevertheless, this may not be always met under the practical application scenarios. Meanwhile, the training procedure of DNN via back-propagation is computationally expensive in terms of both computational resource and time consumption~\cite{VGG}. And, its convergence is sensitive to parameter setting (e.g., learning rate). Thus, for some lightweight applications (e.g., embedded system) more efficient and stable approach is required.

Concerning the defects of the existing shallow and deep metric learning approaches, a novel ensemble cascade metric learning (ECML) mechanism for face verification is proposed by us. The key research motivation is to \emph{seek the good tradeoff between underfitting and overfitting}. Inspired by the success of deep metric learning framework, hierarchical linear metric learning is executed on the raw feature in the cascade way. It aims to improve the discriminative power of the learnt metric by resisting underfitting. However, like each coin has two sides, this also tends to lead to overfitting on the certain feature dimensions.  To alleviate this, we take advantage of the ensemble learning principle~\cite{ensembling} by randomly splitting the features into non-overlapping groups at each learning stage, besides the final one. Then, metric learning is executed among the feature groups individually. Thus, the discriminative information within the feature dimensions can be better maintained to impair the overfitting effect caused by cascade learning scheme. The main technical pipeline of the proposed ensemble cascade metric learning mechanism is shown in Fig.~\ref{fig:pipeline}.

As aforementioned, effective linear metric learning approach is the essential element of the proposed ECML mechanism. To ensure learning efficiency and scalability, the ones with closed-form solution are preferred.  To this end, KISSME~\cite{KISSme} and XQDA~\cite{XQDA} are the outstanding ones. However, we argue that the robustness of KISSME is not satisfactory enough due to the potential computation failure problem on inversion of covariance matrix for Gaussian model, especially in small-scale cases. The main reason is that, the face samples from the same person are often highly correlated. In XQDA, this problem is averted by adding a small turbulence to the within-class  covariance matrix as regularizer. But this may lead to
inaccurate estimation of the within-class covariance matrix. Towards this, a robust Mahalanobis metric learning (RMML) method is proposed by us. It does not need to compute inverse matrix of intra-class covariance matrix, and is of closed-form solution. Meanwhile, the feature distribution characteristics of face  is emphatically considered in RMML to enhance performance. By embedding RMML into ECML mechanism, our proposed metric learning manner (EC-RMML) can run in one-pass learning manner of high scalability.

Experiments on the large-scale Ms-Celeb-1M dataset~\cite{guo2016ms} demonstrate that EC-RMML is superior to state-of-the-art metric learning methods on face verification. And, ECML mechanism is also applicable to other metric learning approaches.

The main contributions of this paper include:

$\bullet$ ECML: a novel ensemble cascade metric learning mechanism for face verification. It is easy to implement, and able to achieve the good tradeoff between underfitting and overfitting;

$\bullet$ RMML: a robust Mahalanobis metric learning approach with closed-form solution.

The source code and supporting materials of our proposition is available at \url{https://github.com/xf1994/ECML}.

\section{Related work} \label{sec:related_work}
To facilitate face verification, one research avenue is to exploit the discriminative visual features.  The well-established ones include LBP and its variants~\cite{Chen2012Bayesian,Chen2013Blessing,LBP3,LBP4}, face attribute learning~\cite{Kumar2010Attribute}, and Fisher vector encoded dense SIFT~\cite{FVface}. Most recently, deep learning-based (i.e., CNN) features~\cite{DeepID,FaceNet,centerloss} have achieved great success.

On top of these visual features, metric learning technology~\cite{mltrack,clustering1,clustering2} has drawn much attention for further performance enhancement. Davis~\emph{et al.}~\cite{ITML} proposed information theoretic metric learning (ITML) method to minimize the differential relative entropy between two multivariate Gaussians. Guillaumin~\emph{et al.}~\cite{LDML} proposed to learn a metric from  the probabilistic perspective using logistic discriminant, which is termed as LDML. Weinberger~\emph{et al.}~\cite{LMNN} proposed the large-margin nearest neighbor (LMNN) metric learning approach to enhance the performance of KNN classification. All the methods mentioned above suffer from one common defect. That is, they highly depend on the iterative optimization procedure during training. As a consequence, their scalability is not satisfactory enough especially towards large-scale data.

To alleviate this, some scalable metric learning approaches are proposed. Among them, LDA~\cite{LDA} runs efficiently with the closed-form solution. However, it cannot work when only the pair-wised labels are given. To address this, SILD~\cite{SILD} is proposed in the way of estimating within-class covariance matrix and between-class covariance matrix using side-information.  Koestinger \emph{et al.}~\cite{KISSme} proposed KISSME learnt through Gaussian hypothesis to explore feature difference space. With the closed-form solution, KISSME works well on both small and large-scale datasets. Nevertheless, it may fail to work occasionally due to the collapse of computing the inversion of covariance matrix for Gaussian model. Meanwhile, the embedded Gaussian hypothesis may not hold towards the high-dimensional features. Taking advantages of KISSME and LDA, Liao \emph{et al.}~\cite{XQDA} proposed cross-view quadratic discriminant analysis (XQDA) metric learning method. XQDA is also learnt through feature difference. It achieves the closed-form solution using the generalized eigenvalue decomposition. Unfortunately, it suffers from the same computational problem as KISSME. That is, computing the inverse matrix of within-class covariance matrix is also required. Hence, in the implementation of XQDA, it adds a small turbulence to the within-class covariance matrix to guarantee its robustness. However, this may lead to inaccurate estimation of the within-class covariance matrix.

{All the metric learning methods above are of shallow learning paradigm. That is, only one global metric is learned from the raw feature. They may often suffer from the underfitting problem. To obtain more reliable metric, some works~\cite{dong2017dimensionality,paisitkriangkrai2015learning} address this issue by the ensemble of multiple metrics. However, the ensemble of shallow metrics still cannot effectively alleviate the intrinsic underfitting problem.}

Very recently, taking advantage of the strong fitting power of deep neural network, some deep metric learning approaches~\cite{Yi2014Deep,Hoffer2014Deep} are proposed in the end-to-end manner. Nevertheless, they are data-hungry. When the training data is not sufficient enough they do not perform well~\cite{Li2014DeepReID}. Inspired by the success of deep metric learning paradigm, we propose ECML as an ensemble cascade metric learning mechanism in spirit of balancing effectiveness and efficiency. Being different from the deep metric learning methods in~\cite{Yi2014Deep,Hoffer2014Deep}, ECML can run in one-pass learning manner without iterative training procedure. To alleviate over-fitting risk, ECML executes feature shuffle operation~\cite{xiangyu2017shufflenet}. Meanwhile, a robust Mahalanobis metric learning method with the closed-form solution is also proposed as the basic unit of ECML. It does not need to compute the inverse matrix of within-class covariance matrix to avoid the computation failure problem faced by KISSME.



\section{ECML: a novel ensemble cascade metric learning mechanism}
\begin{figure}[t]
	\centering
    \footnotesize
	\begin{minipage}{0.46\linewidth}
		\centerline{\includegraphics[width=4.2cm]{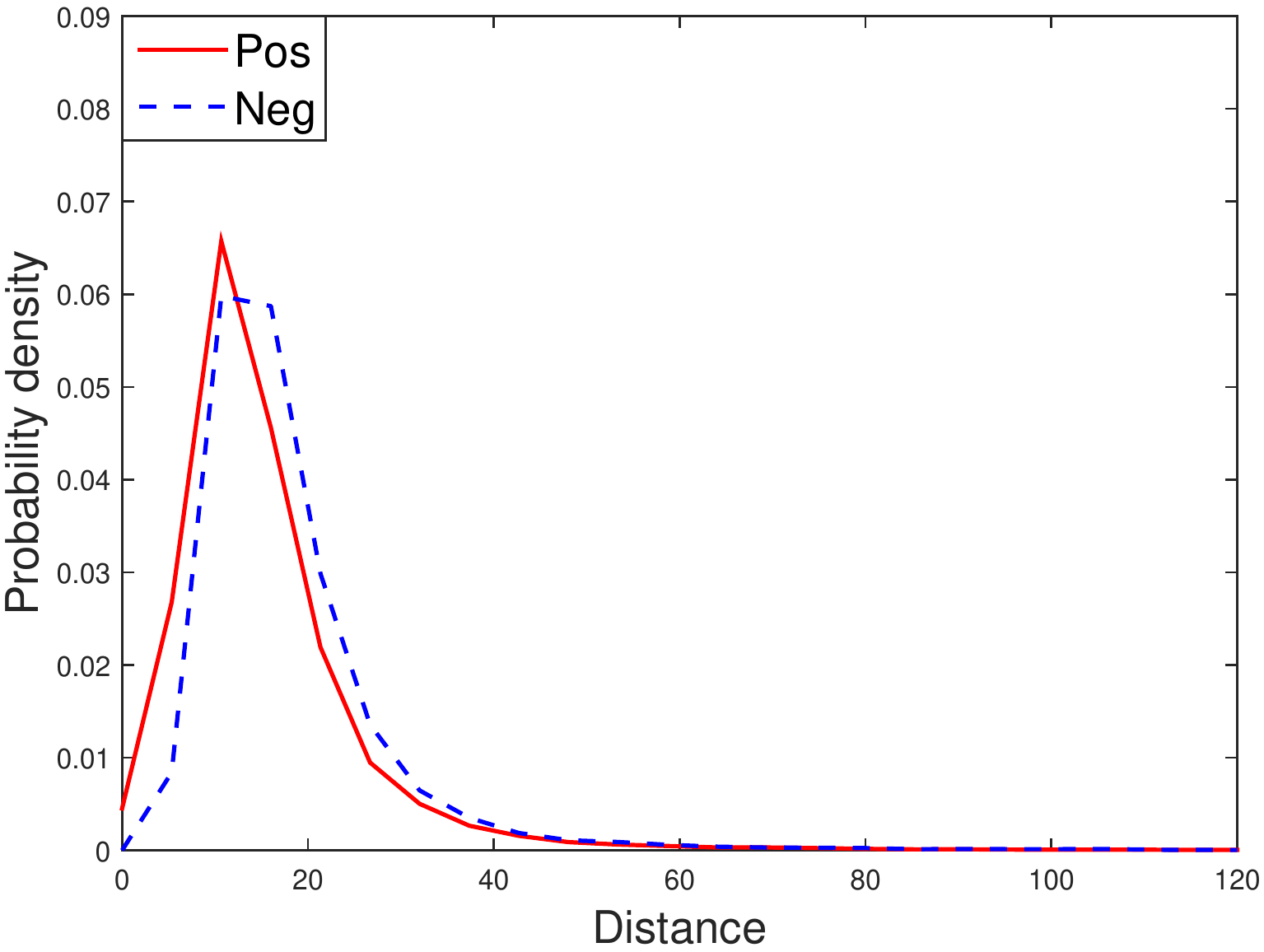}}
		\centerline{(a) Raw distribution}
	\end{minipage}
	\begin{minipage}{0.5\linewidth}
		\centerline{\includegraphics[width=4.2cm]{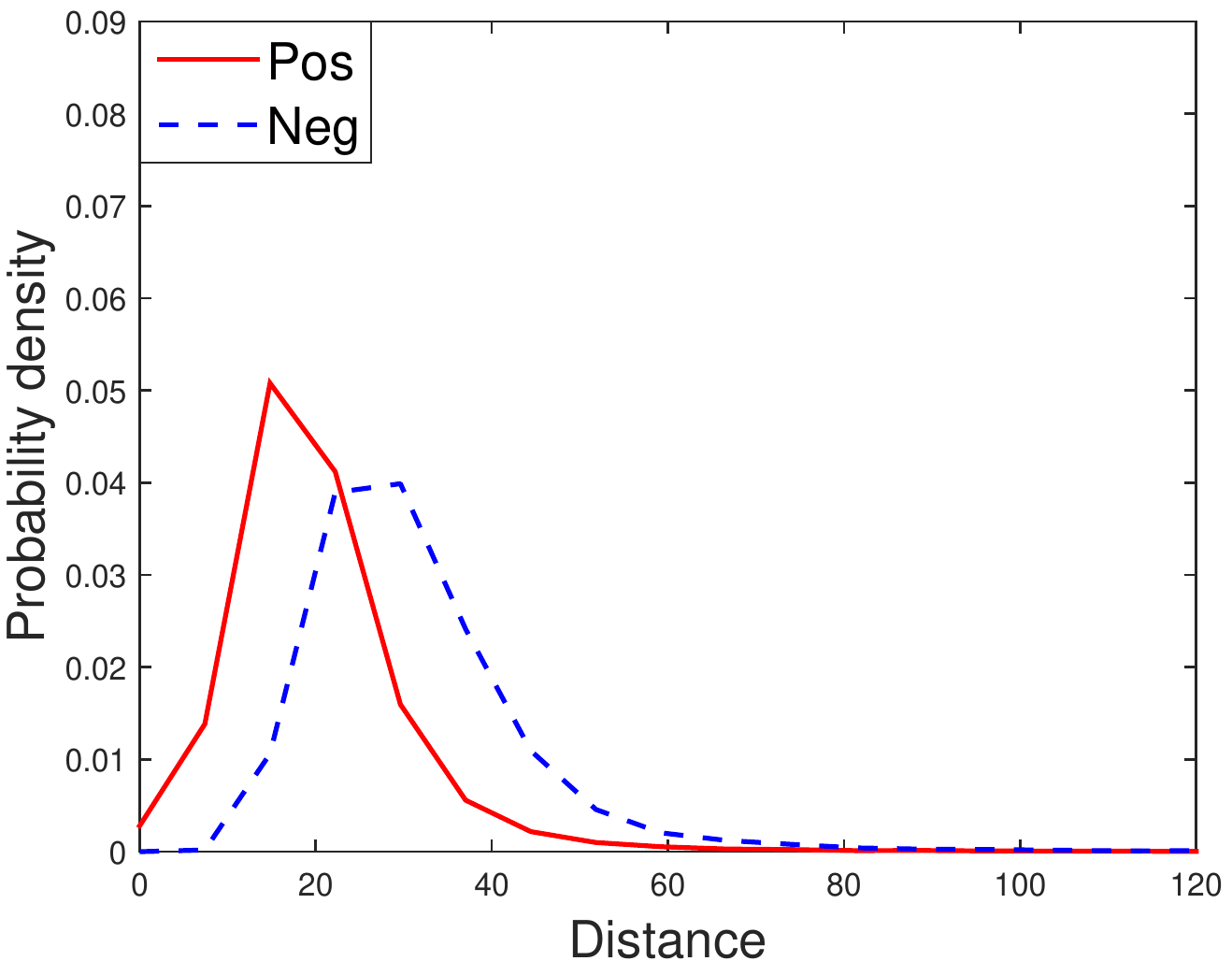}}
		\centerline{(b) With XQDA}
	\end{minipage}
	\caption{Pairwise distance distribution between the subjects within the 55,043 matched face pairs and 44,957 unmatched face pairs randomly sampled from MS-Celeb-1M dataset~\cite{guo2016ms}, using FV-based face representation~\cite{FVface}. \textbf{``Pos"} and \textbf{``Neg"} indicate the matched and unmatched case respectively. Particularly, (a) shows the raw pairwise distance distribution, and (b) shows the pairwise distance distribution after XQDA is executed.}
	\label{fig:kissme_xqda_dist}
\end{figure}

As aforementioned, face verification is indeed a challenging fine-grained visual recognition task. One main difficulty is that, the different people may be of subtle feature difference. To verify this, Fig.~\ref{fig:kissme_xqda_dist} (a) shows the pairwise distance distribution between the subjects within the 100,000 randomly sampled matched and unmatched face pairs (55,043 matched pairs, and 44,957 unmatched pairs) from MS-Celeb-1M dataset~\cite{guo2016ms} using FV-based face representation~\cite{FVface}. It can be observed that the subject distance of the matched and unmatched face pairs distributes have serious overlap, which actually imposes great challenge for accurate face verification. As a consequence, metric learning approach of strong fitting capacity is required to facilitate the discriminative power of feature. Nevertheless, resisting the overfitting risk should be taken into consideration simultaneously. In spirit of achieving good tradeoff between underfitting and overfitting, we propose an ensemble cascade metric learning (ECML) mechanism towards face verification. That is, hierarchical metric learning procedure is executed in cascade way to enhance the fitting power. And, at each learning stage the feature is randomly shuffled into groups to work in ensemble manner to alleviate overfitting.

\subsection{Cascade metric learning to enhance discriminative power} \label{sec:cascade}

Generally, the existing state-of-the-art linear metric learning approaches (e.g., XQDA~\cite{XQDA}) share one common characteristics that they are executed in shallow learning manner (i.e., one-stage learning). We argue that, this paradigm somewhat limits the fitting capacity of metric learning procedure for face verification, being trapped in underfitting status. To reveal this point intuitively, Fig.~\ref{fig:kissme_xqda_dist} (b) shows the pairwise distance distribution between the subjects within the 100,000 face pairs in Fig.~\ref{fig:kissme_xqda_dist} (a) after using XQDA. We can see that, although with the promotion of XQDA the subject distance distribution that corresponds to the matched and unmatched face pairs is still of high overlap. That is to say, underfitting phenomenon happens. Hence, the fitting capacity of the employed metric learning approaches should be further facilitated to improve the discriminative power. As a consequence, we propose to execute cascade metric learning procedure to address this, which is inspired by the great success of deep learning framework that conducts hierarchical nonlinear feature learning ~\cite{VGG}.

\begin{figure}[t]
    \centering
    \includegraphics[width=8.8cm]{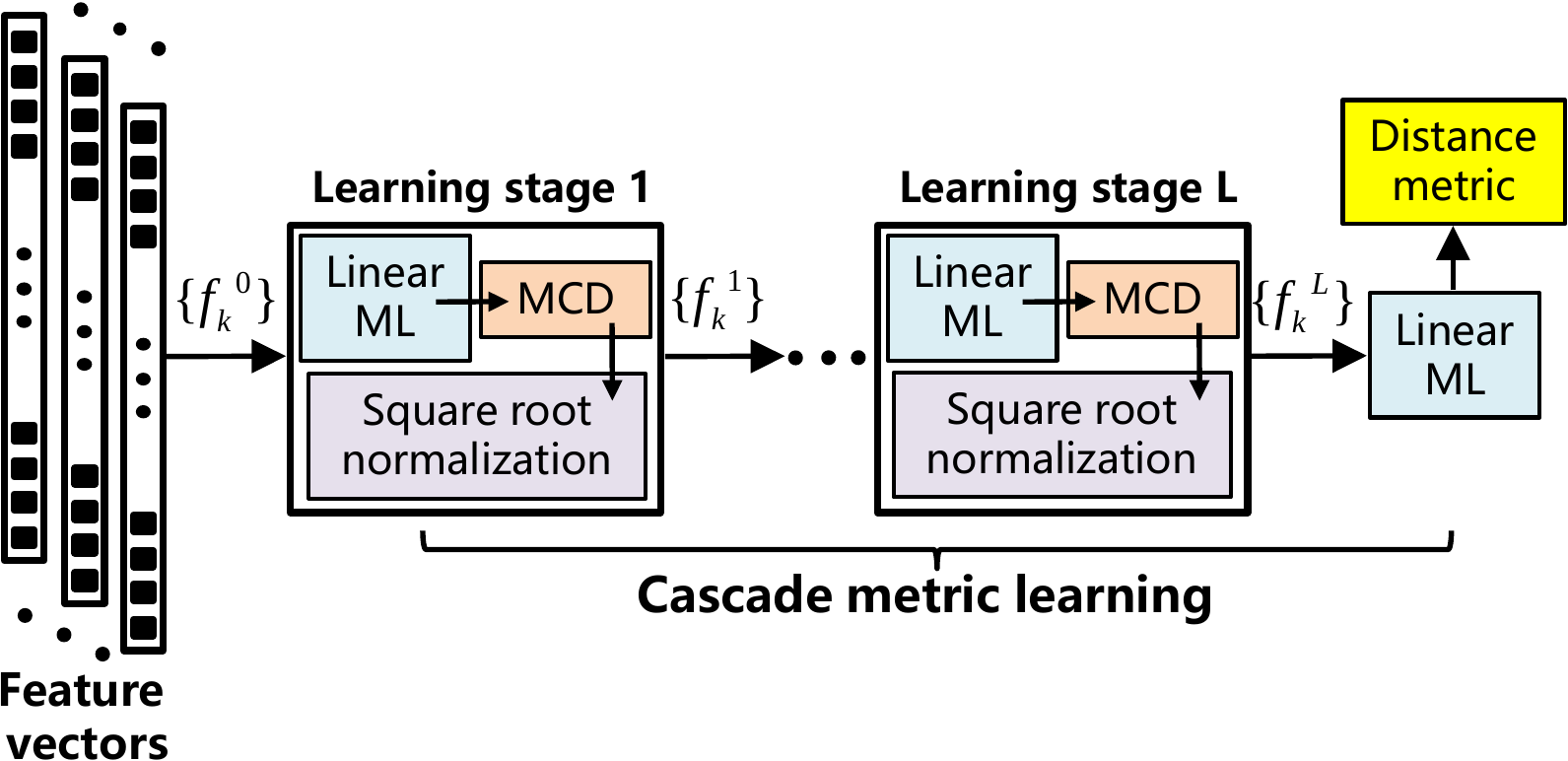}
    \caption{The main technical pipeline of the proposed cascade metric learning mechanism. In particular, \textbf{``ML"} indicates metric learning, \textbf{``MCD"} denotes the modified Cholesky decomposition, and \textbf{${``f_k^l"}$} represents the yielded feature that corresponds to the $k$-th person at the $l$-th learning stage.}
     \label{fig:cascade}
 \end{figure}

As shown in Fig.~\ref{fig:cascade}, the proposed cascade metric learning manner consists of $L+1$ hierarchical learning stages. Our intrinsic intuition is that, \emph{when the metric learning stage goes further, the same person can continuously approach closer while the different persons will be pushed farther} in feature space to alleviate the problem of underfitting. In particular, the first $L$ learning stages will map the face feature from the previous stage to the new feature space. And, the last learning stage yields the final distance metric for face verification.

Linear metric learning procedure is executed in all of the $L+1$ learning stages as the core component, to generate the stage-wise Mahalanobis transformation matrix $M^l$ at the $l$-th learning stage. The main reason for why we choose linear metric learning paradigm to build cascade metric learning mechanism is mainly due to its relatively high computational efficiency and scalability, compared to the nonlinear ones \cite{xiong2014kernelmetric}.

As aforementioned, each of the first $L$ learning stage needs to map the face feature from the previous learning stage to the new face space. Without losing generality, assuming we are at the $l$-th learning stage with the achieved stage-wise Mahalanobis transformation matrix $M^l$. Let $f_k^l$ denote the yielded face feature that corresponds to the $k$-th person at the $l$-th learning stage. One intuitive way to map $f_{k}^{l-1}$ from the previous learning stage is to decompose $M^l$ via Cholesky decomposition as
\begin{equation}
M^l=P^l{P^l}^T.
\label{eq:cholesky_decom}
\end{equation}
Then, $P^l$ is used as the mapping matrix to acquire the output feature of the $l$-th learning stage for the $k$-th person by
\begin{equation}
f_{k}^{l}=P^lf_{k}^{l-1}.
\end{equation}
However, for some existing metric learning approaches (e.g., KISSME) the learnt $M^l$ may not be positive definite. In this case, Cholesky decomposition cannot be executed directly. To address this, we propose a modified Cholesky decomposition approach. In particular, Schur decomposition is first conducted on $M^l$ as
\begin{equation}
M^l=Q^l{\Lambda^l}{Q^l}^T,
\label{eq:schur_decom}
\end{equation}
{where $Q^l$ and $\Lambda^l$ are the decomposed matrices obtained by Schur decomposition. In particular, $Q^l$ is the unitary matrix obtained by Schur decomposition. And, $\Lambda^l$ should be a diagonal matrix since $M^l$ is a real symmetric matrix}.

To make $M^l$ decomposable, $M^l$ is modified as $\widehat{M}^l$ by setting the negative eigenvalues in $\Lambda^l$ as 0. Then, $\widehat{M}^l$ can be decomposed as:
\begin{equation}
\widehat{M}^l=Q^l\widehat{\Lambda^l}{(\widehat{\Lambda^l} Q^l)}^T,
\label{eq:mcd}
\end{equation}
where
\begin{equation}
\widehat{\Lambda^l}_{ij}=\left\{\begin{matrix}
\sqrt{\Lambda^l_{ij}},& \Lambda^l_{ij}\geq0\\
 0,&\Lambda^l_{ij}<0
\end{matrix}\right. \,.
\end{equation}
In other words, $\widehat{M}^l$ is decomposed as
\begin{equation}
\widehat{M}^l=P^l{P^l}^T,
\end{equation}
where
\begin{equation}
P^l=Q^l\widehat{\Lambda^l}.
\end{equation}
We term this decomposition procedure of $\widehat{M}^l$ as modified Cholesky decomposition (MCD). Actually, directly setting negative eigenvalues in $\Lambda^l$ as 0 may hurt the discriminative capacity of metric learning approaches. However, generally the number of negative eigenvalues is relatively small. And, the proposed cascade metric learning procedure help to compensate the defect.

\begin{figure}[t]
	\centering
    \footnotesize
	\begin{minipage}{0.46\linewidth}
		\centerline{\includegraphics[width=4.2cm]{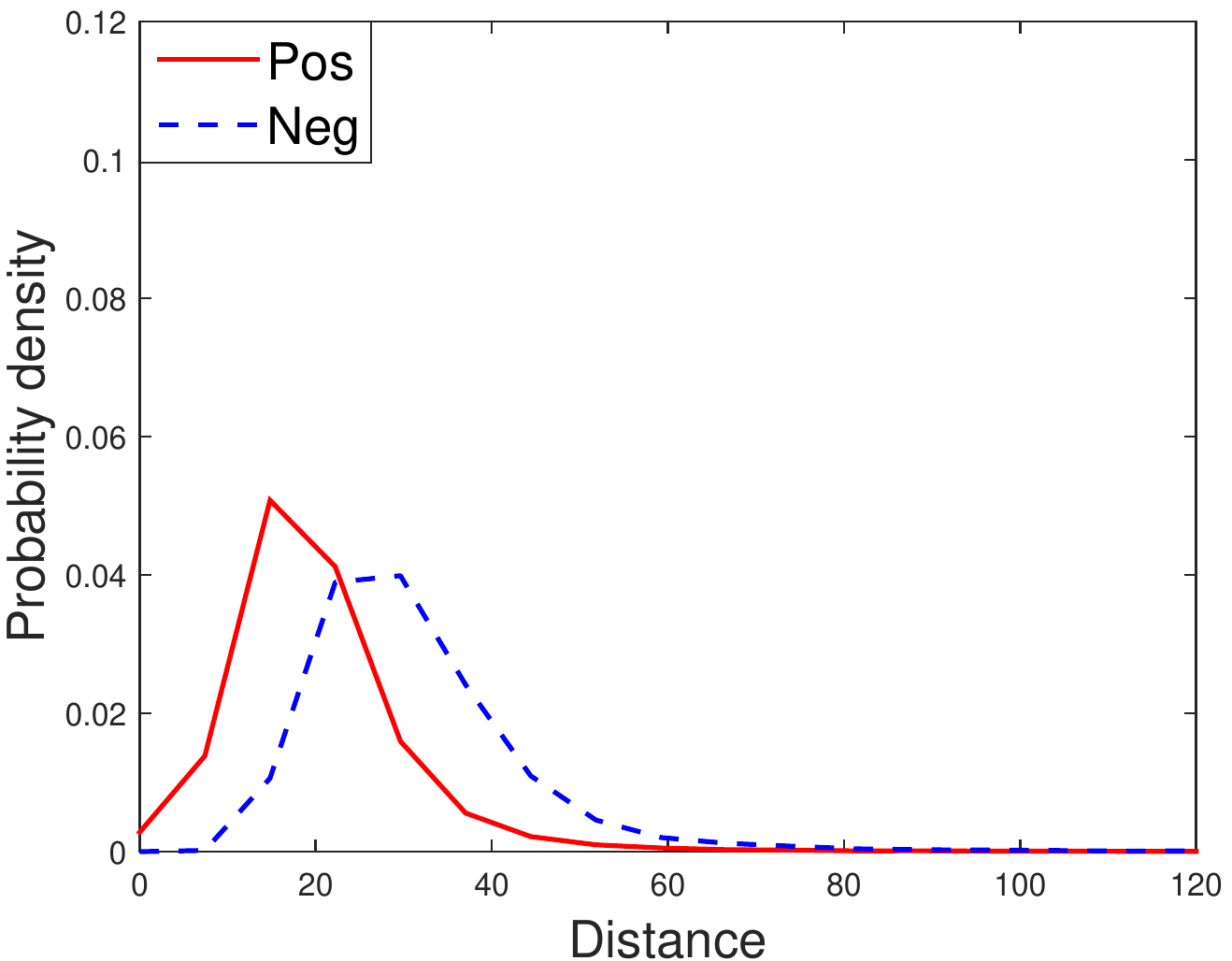}}
		\centerline{(a) Raw XQDA}
	\end{minipage}
	\begin{minipage}{0.5\linewidth}
		\centerline{\includegraphics[width=4.2cm]{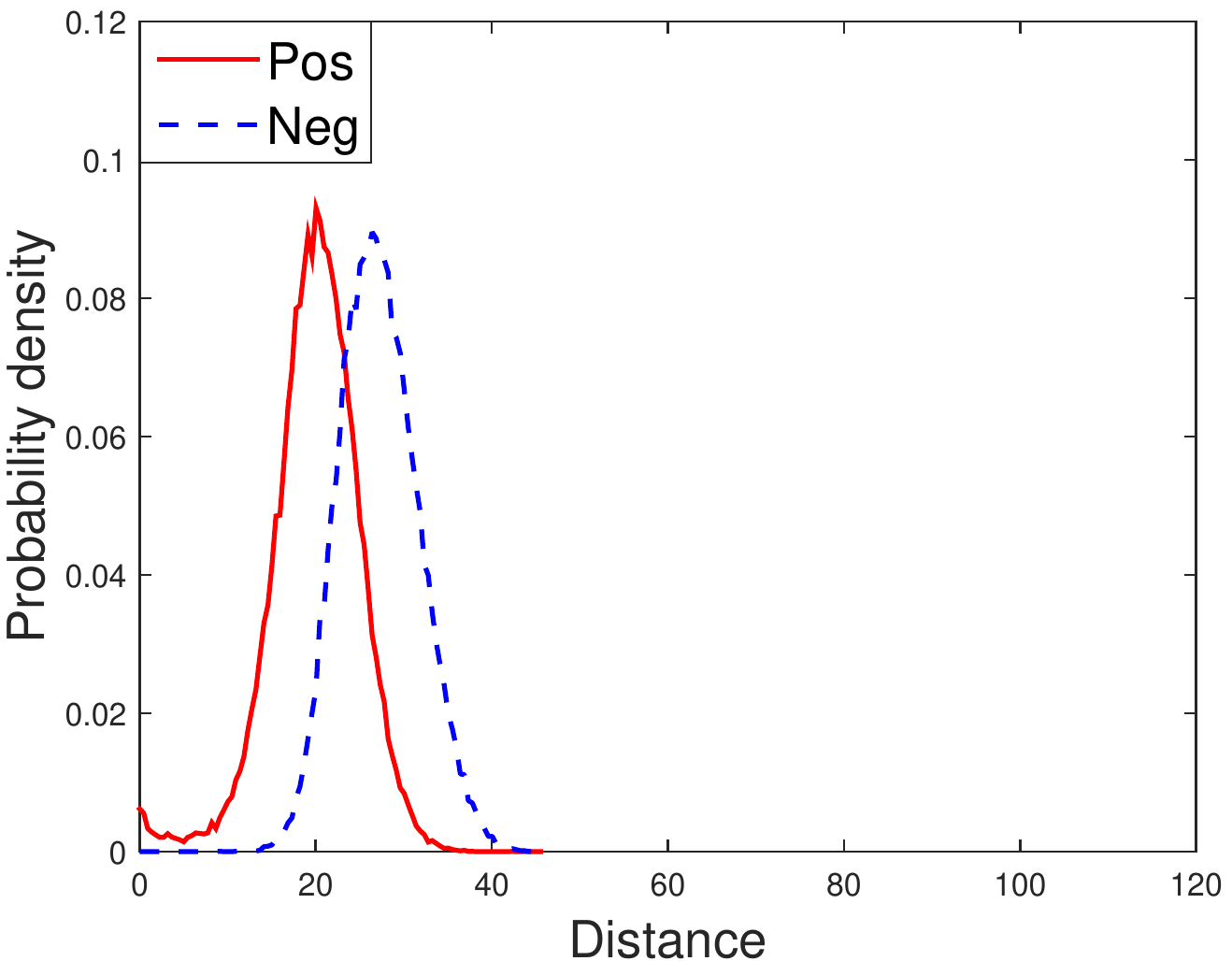}}
		\centerline{(b) Cascade XQDA}
	\end{minipage}
	\caption{Pairwise distance distribution between the subjects within the 55,043 matched and 44,957 unmatched face pairs in Fig.~\ref{fig:kissme_xqda_dist}, using raw XQDA and cascade XQDA respectively. \textbf{``Pos"} and \textbf{``Neg"} indicate the matched and unmatched case respectively. Particularly, (a) shows the distribution of raw XQDA with K-L divergence of 1.2612, and (b) shows the distribution of cascade XQDA with K-L divergence of 1.9142.}
	\label{fig:kissme_raw_cascade}
\end{figure}

Meanwhile, we find that during the phase of cascade metric learning some feature dimensions of the yielded $f_k^l$ may be of much greater values than the other dimensions. Actually, this phenomenon tends to lead overfitting. To alleviate this, we choose to execute square root normalization on $f_k^l$ to suppress the large values as
\begin{equation}
 \varphi\left({f_{k}^{l}}\right)=\sgn\left({f_{k}^{l}}\right)\left|{f_{k}^{l}}\right|^{\frac{1}{2}},
\end{equation}
where $\varphi\left({\cdot}\right)$ indicates the square root normalization operation function; $\sgn\left(\cdot\right)$ denotes the signum function.

To verify the feasibility of cascade metric learning mechanism, we apply it to XQDA. Fig.~\ref{fig:kissme_raw_cascade} shows pairwise distance distribution comparison between the subjects within the face pairs in Fig.~\ref{fig:kissme_xqda_dist} (a), after using raw XQDA and cascade XQDA~\footnote{The metric learning stage number $L$ is empirically set to 3, besides the final learning stage.} respectively. It can be observed that, the subject distance distribution overlap between the matched and unmatched face pairs has been remarkably reduced by cascade XQDA, compared to the raw one. That is, the K-L divergence between the ``Pos" and ``Neg" distribution has been enhanced from 1.2294 to 1.9142.  Although cascade metric learning paradigm is able to enhance the discriminative power of feature on training set, it still may lead to overfitting problem. Next, we will illustrate the way to alleviating this via ensemble metric learning.

 \begin{figure}[t]
    \centering
    \includegraphics[width=8.8cm]{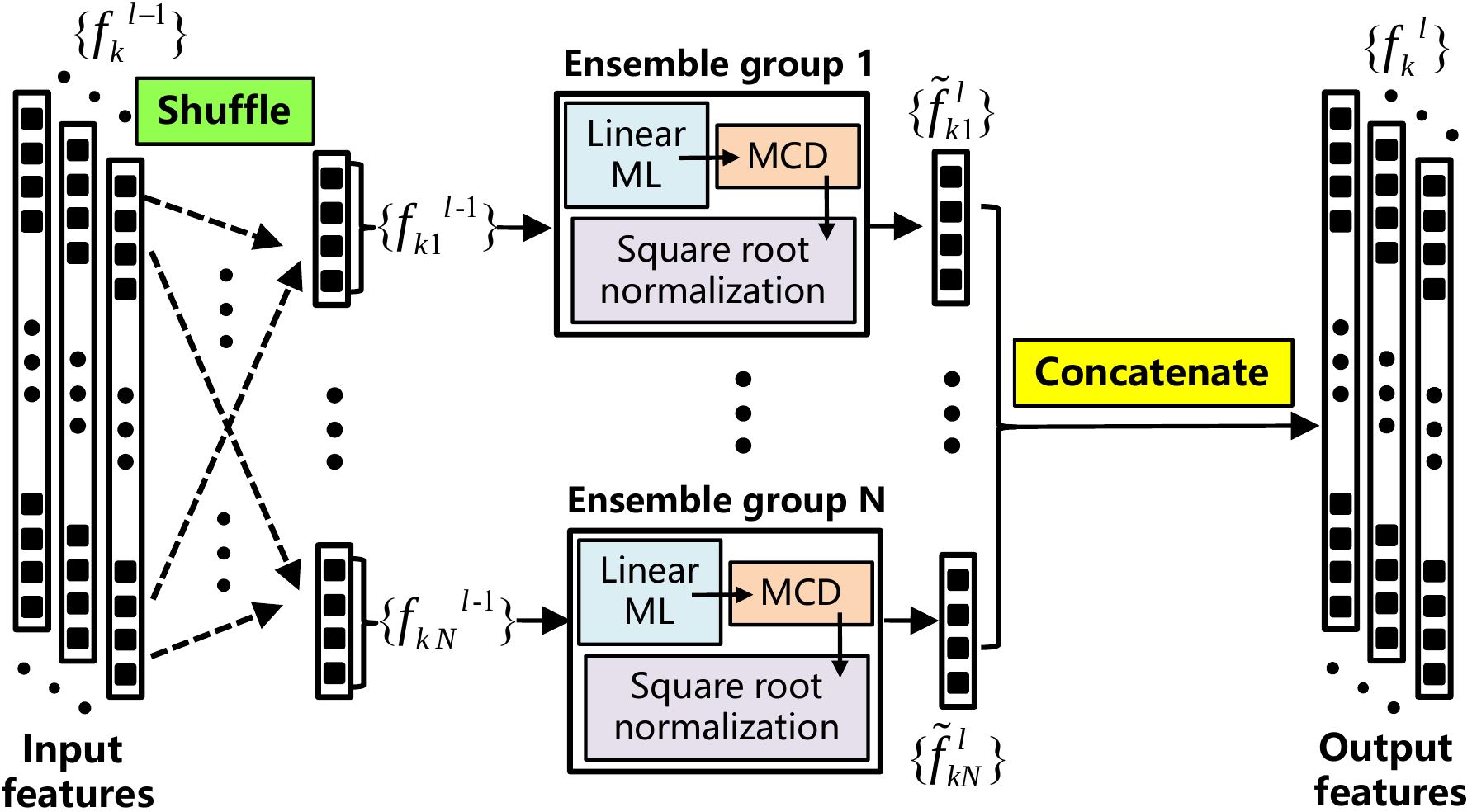}
    \caption{The main technical pipeline of the proposed ensemble metric learning mechanism at the $l$-th cascade learning stage.}
     \label{fig:ensemble}
 \end{figure}

 \subsection{Ensemble metric learning to suppress overfitting risk}

 As verified in Fig.~\ref{fig:kissme_raw_cascade}, cascade metric learning helps to improve the discriminative power of the yielded feature on training set. However, it cannot ensure the generalization capacity to test set. That is, overfitting may happen, which is also often faced by deep learning paradigm~\cite{Dropout}. To suppress overfitting, an ensemble metric learning mechanism is proposed here. Our main idea is that, at each cascade metric learning stage (besides the final one) each input feature vector will be randomly shuffled into groups. Then, metric learning procedure that involves linear metric learning, MCD, and square root normalization proposed in Sec.~\ref{sec:cascade} will be executed to each feature group in ensemble manner. The outputs from all of the ensemble groups are consequently concatenated to yield the final output feature of the corresponding learning stage. Actually, \emph{this can avoid the problem that only a small number of feature dimensions play the dominant role during the phase of metric learning}.  In this way, the discriminative information within the different feature dimensions can be better maintained, which helps to suppress overfitting.

The main technical pipeline of the proposed ensemble metric learning mechanism at the $l$-th learning stage is shown in Fig~\ref{fig:ensemble}. In particular, at the $l$-th cascade learning stage suppose the dimensionality of the input feature vector $f_k^{l-1}$ from the previous learning stage is $h$, we randomly shuffle $f_k^{l-1}$ into $N$ ensemble groups uniformly. That is, each feature group is of dimensionality $D=h/N$. If $h$ cannot be divided by $N$, 0 padding on $f_k^{l-1}$ is executed. Within each group, metric learning procedure is executed with the output feature vector $\tilde{f}_{km}^{l}$, where $m$ is the ensemble group index. Then, the output feature $f_k^l$ of the $l$-th cascade learning stage will be the concatenation of all $\tilde{f}_{km}^{l}$ as $f_k^l = \left[\tilde{f}_{k1}^{l},...\tilde{f}_{km}^{l},...\tilde{f}_{kN}^{l}\right]$.

When the cascade metric learning procedure goes further, the number of the stage-wise ensemble groups will be gradually decreased. Suppose $L$ ensemble metric learning stages exist, the number of the ensemble groups at the $l$-th learning stage is set to $N_l=2^{L-l+1}$ (i.e., $\frac{1}{2}$ as the previous stage). In this way, the ensemble learning procedure gradually fuses the yielded weak metrics to generate the stronger one. Being combined with cascade metric learning, ensemble metric learning aims to alleviate the potential overfitting problem. Using the 100,000 matched and unmatched face pairs in Fig.~\ref{fig:kissme_xqda_dist} as training set, 10,000 face pairs (4,977 matched, and 5,023 unmatched) are additionally randomly sampled from  MS-Celeb-1M dataset as test set to verify the generalization capacity of ensemble metric learning. The comparison on K-L divergence between the ``Pos" and ``Neg" pairwise distance distribution on training and test set among the different metric learning methods is listed in Table~\ref{tab:kl_comp}. XQDA and RMML (proposed in Sec.~\ref{sec:rmml}) are employed as the basic metric learning approaches. Cascade learning procedure and ensemble cascaded learning procedure are imposed to them respectively. It can be observed that:

$\bullet$ For XQDA and RMML, ensemble metric learning can further enhance K-L divergence on test set when it is appended to cascade metric learning;

$\bullet$ Ensemble cascade metric learning mechanism can significantly enlarge K-L divergence on test set, compared to the raw FV feature and basic metric learning approaches.

The results above somewhat verify our propositions that, (1) ensemble metric learning indeed helps to suppress overfitting risk, and (2) ensemble cascade metric learning is essentially effective to improve the discriminative power of raw feature.

\section{RMML: a robust Mahalanobis metric learning approach} \label{sec:rmml}
\subsection{Revisit on KISSME}
As we can see from Fig.~\ref{fig:cascade} and~\ref{fig:ensemble}, Mahalanobis (i.e., linear) metric learning approach plays the fundamental role within the proposed ECML mechanism. An effective and efficient Mahalanobis metric learning approach is preferred to drive ECML. Among the existing linear ones, KISSME~\cite{KISSme} with the closed-form solution is widely used due to its balance between effectiveness and efficiency. Here, we will make a revisit on KISSME first.

Let $X=\left\{x_1,x_2...x_N\right\}$ denote the training sample set, $d_{ij}=x_i -x_j$ represent the feature difference between 2 samples, and $y_{ij}$ indicate which indicate whether $x_i$ and $x_j$ belong to the same person ($y_{ij}=1$) or not ($y_{ij}=0$). KISSME assumes a zero-mean Gaussian structure of the feature difference space. The likelihood towards whether $\left(x_i, x_j\right)$ corresponds to the matched pair or not can be estimated as
\begin{equation}
P(d_{ij}|H_{Pos})=\frac{1}{(2\pi)^{D/2}|\Sigma_{Pos}|^{1/2}}e^{-\frac{1}{2}d_{ij}^T\Sigma_{Pos}^{-1}d_{ij}},
\end{equation}
and
\begin{equation}
P(d_{ij}|H_{Neg})=\frac{1}{(2\pi)^{D/2}|\Sigma_{Neg}|^{1/2}}e^{-\frac{1}{2}d_{ij}^T\Sigma_{Neg}^{-1}d_{ij}},
\end{equation}
where $D$ indicates the feature vector dimensionality; $H_{Pos}$ denotes the hypothesis that $\left(x_i, x_j\right)$ is the matched pair, and $H_{Neg}$ represents the hypothesis that $\left(x_i, x_j\right)$ is the unmatched pair. And, $\Sigma_{Pos}$ and $\Sigma_{Neg}$ are covariance matrices for the matched and unmatched cases respectively. They are computed as
\begin{equation}
\Sigma_{Pos}=\sum_{y_{ij}=1}d_{ij}d_{ij}^T,
\end{equation}
and
\begin{equation}
\Sigma_{Neg}=\sum_{y_{ij}=0}d_{ij}d_{ij}^T.
\end{equation}
By applying the log-likelihood ratio test, the distance function can be simplified as
\begin{equation}
\label{eqn:kiss}
f(d_{ij})=d_{ij}^T(\Sigma_{Pos}^{-1}-\Sigma_{Neg}^{-1})d_{ij}.
\end{equation}


\begin{table}[t]
\small
\caption{Comparison on K-L divergence between the ``Pos" and ``Neg" pairwise distance distribution on training and test set, among the different metric learning methods. \textbf{``ML"}, \textbf{``C-"} and \textbf{``EC-"} indicate metric learning, cascade learning mechanism, and ensemble cascade learning mechanism respectively.}
\centering
\begin{tabular}{|c|c|c|}
\hline ML method &  Training set & Test set \\ \hline \hline
Raw FV feature & 0.2209 & 0.1655\\ \hline \hline
XQDA & 1.2294 & 1.2612\\ \hline
C-XQDA & 1.9142 & 2.0720\\ \hline
EC-XQDA & 1.7940 & 2.2295\\ \hline \hline
RMML & 0.2561 & 0.1773\\ \hline
C-RMML & 0.9292 & 0.9539\\ \hline
EC-RMML & 1.0643 & 1.1246\\ \hline
\end{tabular}
\label{tab:kl_comp}
\end{table}

\begin{figure}
\centering
\includegraphics[width=7cm]{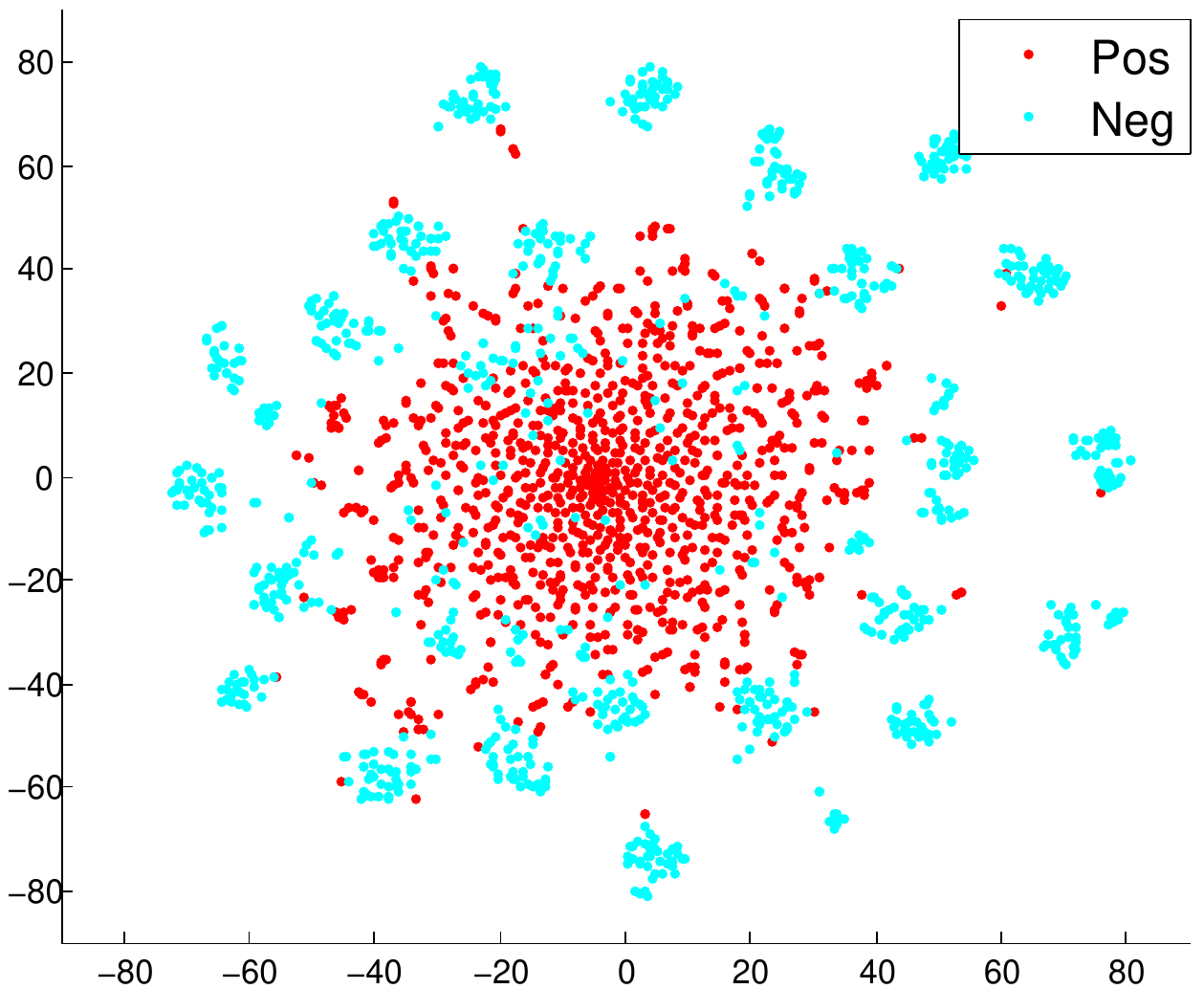}
\caption{Feature difference distribution of the 2,000 matched and unmatched face pairs using CNN feature. \textbf{``Pos"} denotes the matched case, and \textbf{``Neg"} indicates the unmatched case. It is drawn using t-SNE{\cite{tsne}}.}
\label{fig:RMML}
\end{figure}

KISSME is indeed an effective and efficient linear metric learning method. However, we argue that it still suffers from 2 main defects, according to the revisit above as follows:

$\bullet$  First, it needs to fit Gaussian models for the feature difference (not unsigned distance) that correspond to the matched and unmatched sample pairs. However towards face, the feature difference does not always obey Gaussian distribution. To verify this, Fig.~\ref{fig:RMML} shows the feature difference distribution of the 100,000 matched and unmatched face pairs in Fig.~\ref{fig:kissme_xqda_dist} when using CNN feature. It can be clearly observed that, the feature difference distribution of the unmatched face pairs does not distribute in Gaussian form;

$\bullet$ Secondly, KISSME needs to compute the inversion of covariance matrices towards the Gaussian models that correspond to the feature difference on match and unmatched face pairs ($\Sigma_{Pos}^{-1}$ for matched case, and $\Sigma_{Neg}^{-1}$ for unmatched case). Nevertheless since the matched face pairs are always highly correlated as shown in Fig.~\ref{fig:RMML}, $\Sigma_{Pos}$ tends to be singular in cases. Thus, the computation on $\Sigma_{Pos}^{-1}$ may fail to work occasionally. Although PCA is applied to alleviate this problem to some extent in KISSME, the execution of PCA may lead to discriminative information loss.

To address the problems in KISSME above for improvement, a robust Mahalanobis metric learning approach (RMML) is proposed by us. It aims to push the feature difference of the matched face pairs close to origin, and pull the feature difference of the unmatched face pairs far from origin to enhance discriminative power. Being difference from KISSME, computation on the inversion of covariance matrices for Gaussian models is not required to enhance robustness. Meanwhile, RMML is of closed-form solution to ensure that it can run in one-pass learning manner without complex iterative optimization procedure.

\subsection{RMML formulation}

Denote $X=\left\{x_1,x_2...x_N\right\}$ as the training set of face samples, following KISSME we cast RMML into the feature difference space as $d_{ij}=x_i-x_j$. Let $y_{ij}$ indicate whether $x_i$ and $x_j$ belong to the same person ($y_{ij}=1$) or not ($y_{ij}=0$). As aforementioned, RMML is proposed to push the feature difference of the matched face pairs close to origin, and pull the feature difference of the unmatched face pairs far from origin. To this end, we propose to search a Mahalanobis matrix $M$ able to minimize intra-class distance and maximize inter-class distance. As a consequence, the discriminative term of RMML is defined as
\begin{equation}
g_1={ \frac{\sum_{y_{ij}=1}d_{ij}^TMd_{ij}}{\sum_{y_{ij}=1}d_{ij}^Td_{ij}}-\frac{\sum_{y_{ij}=0}d_{ij}^TMd_{ij}}{\sum_{y_{ij}=0}d_{ij}^Td_{ij}}},
\end{equation}
{where $\sum_{y_{ij}=1}d_{ij}^Td_{ij}$ and $\sum_{y_{ij}=0}d_{ij}^Td_{ij}$ are used to normalize the distance of the matched and unmatched face pairs to be of comparable magnitude}. Besides the discriminative term, a regularization term is also given by
\begin{equation}
g_2=\frac{1}{2}\|M-I\|_F,
\end{equation}
where $I$ indicates the identity matrix; and $\|\cdot\|_F$ represents Frobenius norm of matrix. The regularization term $g_2$ aims to resist feature space distortion. By incorporating $g_1$ and $g_2$, the learning procedure of RMML is formulated by
    \begin{equation}
    {\hat{M}=\arg\min\limits_M {\lambda}g_1+g_2}
    \label{eqn:rmml}
\end{equation}
where $\lambda$ plays the role to balance the effect of $g_1$ and $g_2$. It is worthy noting that, within RMML we do not require the feature difference distributes in Gaussian form as KISSME.

\subsection{Closed-form solution of RMML}

Eqn.~\ref{eqn:rmml} is actually a convex optimization problem. It can be solved using the existing optimization packages. However, this procedure is time consuming and requires powerful computational resource, especially for large-scale learning tasks. To address this, we propose a closed-form solution of RMML. To solve Eqn.~\ref{eqn:rmml}, the discriminative term $g_1$ is rewritten as
 \begin{equation}
{g_1=\frac{tr(A^TMA)}{tr(A^TA)}-\frac{tr(B^TMB)}{tr(B^TB)},}
 \end{equation}
 where $A$ of size $D\times N_1$ is feature difference matrix for the matched face pairs; $B$ of size $D\times N_2$ is feature difference matrix for the unmatched face pairs; $D$ is the feature dimension number; and $tr\left(\cdot\right)$ indicates the trace of a matrix. In particular, the column of $A$ corresponds to the feature difference of the samples from the same person (i.e., $d_{ij}|y_{ij}=1$), and the column of $B$ corresponds to the feature difference of the samples from the different persons (i.e., $d_{ij}|y_{ij}=0$). Then taking the derivative of ${\lambda}g_1+g_2$ on $M$, we have
 \begin{equation}
 {\nabla_M=\lambda\left(\frac{AA^T}{tr\left(A^TA\right)}-\frac{BB^T}{tr\left(B^TB\right)}\right)+\left(M-I\right)}.
 \end{equation}
 Setting $\nabla_M=0$, it can be obtained that
 \begin{equation}
M=I+{\lambda}\left(\frac{BB^T}{tr\left(B^TB\right)}-\frac{AA^T}{tr\left(A^TA\right)}\right).
 \label{eqn:derivative}
 \end{equation}
In order to make $I$ and $\frac{BB^T}{tr\left(BB^T\right)}-\frac{AA^T}{tr\left(AA^T\right)}$ to be comparable on magnitude, we normalize $\frac{BB^T}{tr\left(BB^T\right)}-\frac{AA^T}{tr\left(AA^T\right)}$ by the mean of its eigenvalues as
\begin{equation}
\widetilde{M}=\frac{1}{\rho}\left(\frac{BB^T}{tr\left(BB^T\right)}-\frac{AA^T}{tr\left(AA^T\right)}\right),
\end{equation}
where $\rho$ is the mean of eigenvalues. Finally, we obtain the closed-form solution of RMML as
\begin{equation}
\hat{M}=I+\lambda\widetilde{M}.
\end{equation}

It is worthy noting that, no computation on inversion of matrix is required to obtain $\hat{M}$. The computation failure problem on $\Sigma_{Pos}^{-1}$ in KISSME will not happen to RMML. Compared to KISSME, RMML is indeed more robust.

\section{Experiments}

\begin{figure}
\centering
\includegraphics[width=7cm]{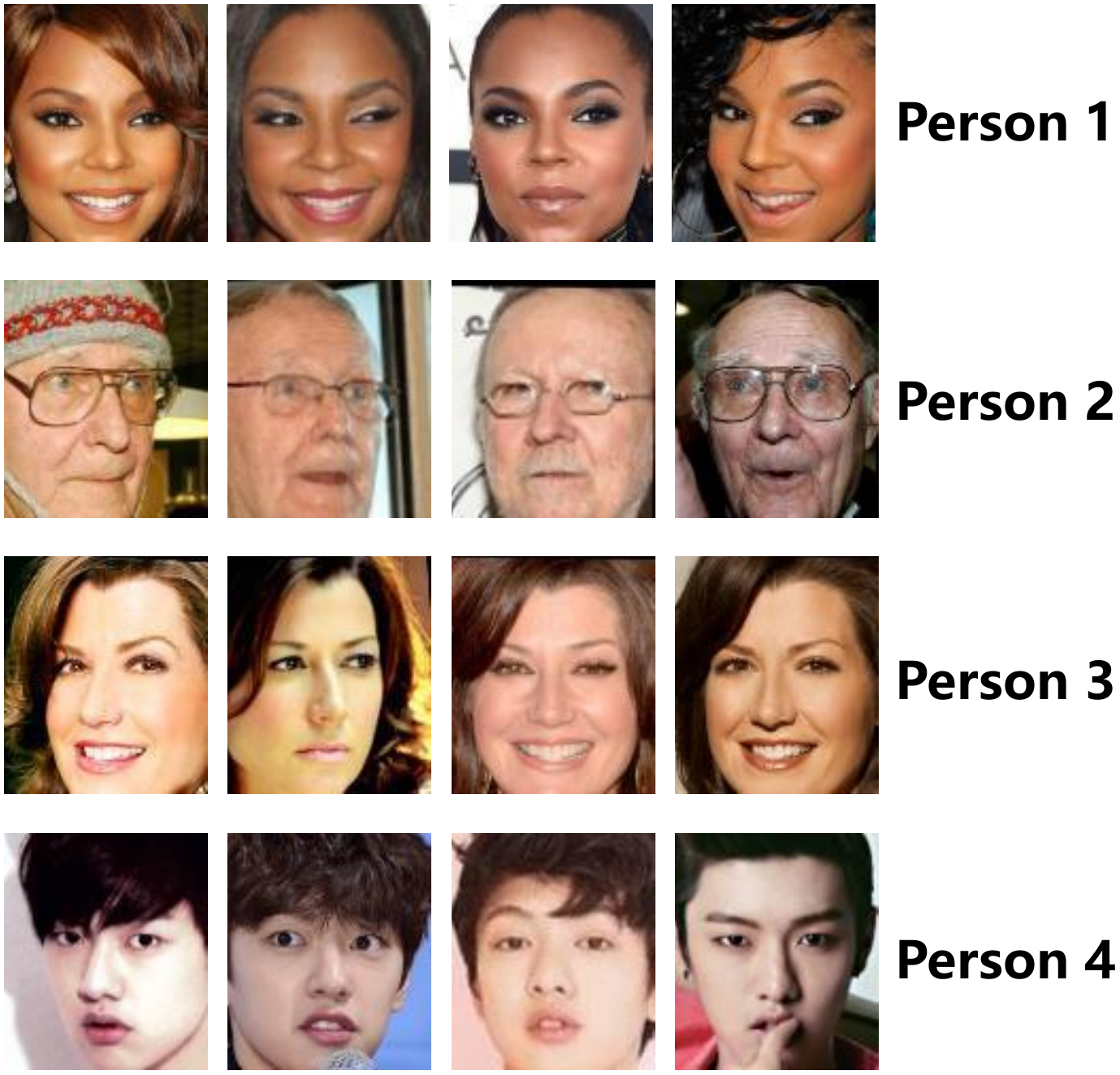}
\caption{Face samples from the MS-Celeb-1M dataset, including male and female. They are of high intra-person variation due to the issues of pose, age, makeup, facial expression, etc.}
\label{fig:faceimages}
\end{figure}

To verify the effectiveness and efficiency of the proposed ensemble cascade metric learning (ECML) mechanism and robust Mahalanobis metric learning approach (RMML), we choose to test them on the recently proposed MS-Celeb-1M dataset~\cite{guo2016ms}. This dataset consists of 1 million celebrities, and each person is of over 20 labeled images. The samples are of high intra-person variation as shown in Fig.~\ref{fig:faceimages}, due to the issues of pose, age, makeup, facial expression, etc. This actually imposes great challenges to accurate face verification. During experiments, all the face images will be regularized to $76\times76$ pixels using the face alignment approach in~\cite{seetaalign}.

To justify the generalization capacity of our propositions, the widely-used deep convolutional neural network (CNN)~\cite{VGG} and Fisher vector (FV)~\cite{FVface} face characterization manners are employed as the input feature of ECML and RMML respectively.

\begin{figure*}
\centering
\includegraphics[width=17cm]{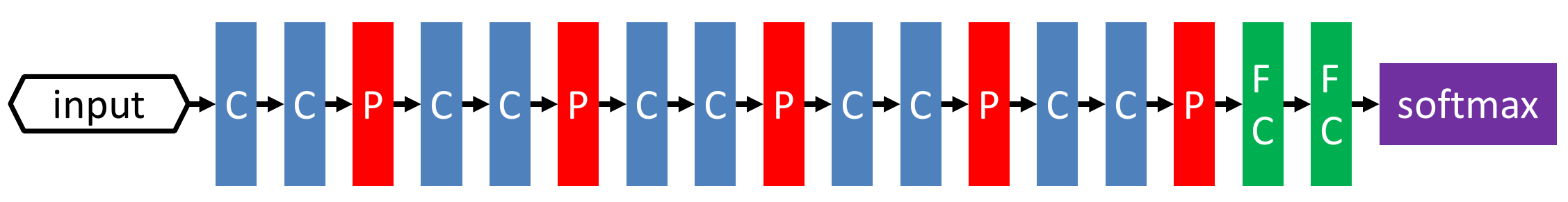}
\caption{The CNN architecture used for face verification. The input image is of size $76\times76$ pixels. \textbf{``C"} indicates the convolution layers using $3\times3$ kernel, with RELU nonlinear units. \textbf{``P"} denotes the max pooling layers with stride 2. \textbf{``FC"} represents the fully-connected layer. The output dimensionality of the first fully connected layer is 640.}
\label{fig:CNNarch}
\end{figure*}

To demonstrate the scalability and robustness of our approach, the experiments are conducted using the small-scale and large-scale test protocols on CNN feature as follows:

$\bullet$ \emph{Small-scale test protocol},  6,000 images are randomly selected from 84 persons for training, and 4,000 images are randomly selected from 52 persons for test. 6,000 training pairs are randomly selected, including 3,169 positive pairs and 2,831 negative pairs. 4,000 verification pairs are randomly selected for test, including  2,160 positive pairs and 1,840 negative pairs;

$\bullet$ \emph{Large-scale test protocol}, 450,000 images are randomly selected from 6,700 persons for training, and 50,000 images are randomly selected from  807 persons for test. 450,000 training pairs are randomly selected, including 218,535 positive pairs and 231,465 negative pairs. 50,000 verification pairs are randomly selected for test, including 26,851 positive pairs and  23,149 negative pairs.

Since the raw FV face feature requires a huge number of computational memory, the experiments are conducted on it using moderate-scale protocol as follow:

$\bullet$ \emph{Moderate-scale test protocol}, 100,000 images are randomly selected from 1267 persons for training, and 10,000 images are randomly selected from 138 persons for test. 100,000 training pairs are randomly selected, including 55,043 positive pairs and 44,957 negative pairs. 10,000 verification pairs are randomly selected for test, including 4,977 positive pairs and 5,023 negative pairs.

Wide-range comparisons with the state-of-the-art metric learning approaches are executed, including LMNN~\cite{LMNN}, LDML~\cite{LDML}, ITML~\cite{ITML}, KISSme~\cite{KISSme}, SILD~\cite{SILD}, and XQDA~\cite{XQDA}. In addition, Mahalanobis matrix for genuine pairs $M_{y=1}$~\cite{KISSme} is learnt as baseline. Since LMNN, XQDA and LDML  need to specify the identity of each face image, full label information is provided for this two methods. The output feature dimensionality of LMNN and LDML is set as the same as the input feature. Parameters within ITML are sept as the default ones. Following~\cite{XQDA}, for XQDA and SILD dimensions with eigenvalue larger than 1 are preserved. For RMML, when it runs independently the parameter $\lambda$ is set to 0.5.  When it is embed into ECML, $\lambda$ is set to 0.1 for all the learning stages. For ECML, the cascade metric learning stage number $L$ is set to 3, plus 1 final linear metric learning stage.

{The existing studies~\cite{KISSme, XQDA} generally choose to reduce the feature dimensionality to a fixed length empirically. However according to our experience, PCA dimensionality can essentially influence the performance of metric learning approaches. Hence, to conduct a thorough investigation on its effect towards the different metric learning methods, PCA is executed with the different output dimensionalities  (i.e., 640, 320, 160, 80 and 40) to the raw feature}. And, we also intend to compare our proposition with the other metric learning approaches in the different PCA cases to demonstrate the superiority.

Euclidean distance is employed to measure the similarity between the 2 face samples for verification. Equal Error Rate (EER)~\cite{deepface} is reported as the performance evaluation criteria on effectiveness. Comparison on running speed of the different approaches will also be executed towards efficiency.

Since random feature shuffle operation exists within ECML (ensemble metric learning phase specifically), the ECML boosted metric learning approaches will run for 5 times and the average EER is reported as the performance evaluation criteria to suppress the effect of randomness. Actually, this leads to more fair comparison among the different metric learning methods.

\subsection{Performance comparison on CNN feature} \label{sec:experiment_cnn}

Deep learning paradigm has brought impressive advance to the state-of-the-art performance on face verification task, CNN especially. Being different from the handcraft features (i.e., SIFT and HOG), CNN possesses the strong capacity of feature learning in data-driven manner for performance enhancement. To conduct experiments on CNN feature, a small CNN network is trained by us based on CASIA dataset~\cite{CASIA}. The employed CNN architecture is shown in Fig.~\ref{fig:CNNarch}. The 640-dimensional output of the first fully-connected layer is employed as face feature. As aforementioned, towards CNN feature experiments are conducted using the small-scale and large-scale test protocols simultaneously. Next, we will introduce the experimental results that correspond to these 2 protocols respectively.

\begin{table}
        \small
        \caption{Performance comparison of EER (\%) among the different metric learning approaches on CNN feature, using \textbf{small-scale} test protocol. \textbf{``EC-"}  represents ensemble cascade learning mechanism. The best performance that corresponds to each PCA dimensionality is shown in boldface.}
        \label{table:CNN_samll}
        \centering
        \begin{tabular}{@{\hspace{2mm}}c@{\hspace{2mm}}c@{\hspace{2mm}}c@{\hspace{2mm}}c@{\hspace{2mm}}c@{\hspace{2mm}}c@{\hspace{2mm}}c@{\hspace{2mm}}}
        \hline
        PCA dim. &640 &320 &160 &80 &40\\
        \hline
        Raw feature  &25.11  &25.00 &24.67 &23.86 &22.61\\
        $M_{y=1}$~\cite{KISSme} &43.10 & 38.53 &34.08 &28.80 &24.58\\
        KISSME~\cite{KISSme} &30.87 &26.79 &25.05 &22.96 &21.74\\
        SILD~\cite{SILD} &38.29 &34.19 &29.40 &25.51 &24.24\\
        LMNN~\cite{LMNN} &23.94 &23.70 &23.29 &22.83 &22.07\\
        ITML~\cite{ITML} &20.44 &20.56 &21.53 &21.69 &22.01\\
        LDML~\cite{LDML} &21.16 &21.56 &19.58 &19.17 &19.08\\
        XQDA~\cite{XQDA} &29.95 &28.37 &25.71 &23.57 &21.79\\
        \hline
        RMML (ours) &25.44 &23.29 &22.01 &20.76 &20.19\\
        EC-RMML (ours) &\textbf{10.64} &\textbf{10.47} &\textbf{10.49} &\textbf{11.55} &\textbf{12.66}\\
        \hline
        EC-KISSME (ours) &-- &-- &-- &-- &--\\
        EC-XQDA (ours) &33.85 &32.60 &30.99 &14.65 &14.93\\
        \hline
        \end{tabular}
\end{table}

\subsubsection{\textbf{Test using small-scale protocol}}

The performance comparison of EER among the different metric learning approaches using thep small-scale test protocol is listed in Table~\ref{table:CNN_samll}. It can be observed that:

$\bullet$ When embedding RMML into ECML mechanism (i.e., EC-RMML), it outperforms the other metric learning approaches significantly in all the test cases that correspond to the different PCA dimensionalities.
That is, the performance enhancement on EER is $32.46\%$ at most and $6.42\%$ at least.
This indeed verifies the effectiveness of our proposition that combines ECML and RMML;

$\bullet$ ECML mechanism is able to enhance the performance of RMML remarkably. The performance gain yielded by ECML is $14.80\%$ at most and $7.53\%$ at least on EER. Hence, this demonstrates that ECML is an effective cascade metric learning mechanism for performance improvement towards face verification;

$\bullet$ Without ECML, RMML achieves comparable performance with the other metric learning methods in the test cases. It is worthy noting that, RMML is derived from KISSME but with better performance consistently. The reason seems that, the feature difference of unmatched face pairs does not distribute in Gaussian form as being assumed by KISSME (demonstrated in Fig.~\ref{fig:RMML}). This also reveals the insight that, the choice of optimal metric learning approach somewhat depends on the raw feature distribution;


$\bullet$ When embedding KISSME into ECML mechanism (i.e., EC-KISSME), it fail to work since the computation failure problem on the inversion of covariance matrices towards the Gaussian models. Nevertheless, it will not happen to RMML. This phenomenon actually demonstrates the robustness of RMML for application. When embedding XQDA into ECML mechanism (i.e., EC-XQDA), it has enhanced the performance of XQDA when PCA dimensionality is set as 80 and 40. This demonstrated the general effectiveness of the ECML mechanism for boosting the performance of linear metric learning methods;

$\bullet$ Generally, all of the involved metric learning approaches can improve the discriminative power of raw feature. And, with the reduction of PCA dimensionality the performance of all the metric learning methods are enhanced in most cases.

\begin{table}
    \small
    \caption{Performance comparison of EER (\%) among the different metric learning approaches on CNN feature, using \textbf{large-scale} test protocol. \textbf{``EC-"}  represents ensemble cascade learning mechanism. The best performance that corresponds to each PCA dimensionality is shown in boldface.}
    \label{table:CNN_large}
    \centering
    \begin{tabular}{@{\hspace{2mm}}c@{\hspace{2mm}}c@{\hspace{2mm}}c@{\hspace{2mm}}c@{\hspace{2mm}}c@{\hspace{2mm}}c@{\hspace{2mm}}c@{\hspace{2mm}}}
    \hline
        PCA dim. &640 &320 &160 &80 &40\\
        \hline
        Raw feature  &23.37  &23.33 &23.00 &22.41 &21.58\\
        $M_{y=1}$~\cite{KISSme} &40.26 & 34.68 &30.24 &25.63 &21.28\\
        KISSME~\cite{KISSme} &22.20 &23.07 &21.56 &19.73 &18.45\\
        SILD~\cite{SILD} &22.04 &24.91 &23.56 &22.44 &21.59\\
        XQDA~\cite{XQDA} &27.07 &25.98 &23.85 &21.32 &18.97\\
        \hline
        RMML (ours) &18.06 &17.68 &17.41 &17.35 &17.72\\
        EC-RMML (ours) &\textbf{12.01} &\textbf{11.43} &11.20 &11.51 &12.35\\
        \hline
        EC-KISSME (ours) &-- &-- &10.14 &10.30 &11.42\\
        EC-XQDA (ours) &31.44 &11.86 &\textbf{9.83} &\textbf{10.09} &\textbf{11.35}\\
\hline
\end{tabular}
\end{table}

\subsubsection{\textbf{Test using large-scale protocol}}

During the phase of large-scale test, since the gradient decent based metric learning approaches are extremely time-consuming we only report the results of the scalable methods of the closed-form or approximate closed-form solution. The performance comparison of EER among the different metric learning approaches using the large-scale test protocol is listed in Table~\ref{table:CNN_large}. From the experimental results, we can see that:

$\bullet$ In the large-scale test case, among all the metric learning approaches EC-RMML still significantly outperforms the other existing ones.
 It is worthy noting that, EC-RMML can work effectively in both large-scale case than small-scale case.
 This actually verifies the generalization capacity of EC-RMML to data scale. Accordingly, we assume that EC-RMML possesses strong potentiality to explore the discriminative information within the big data;

$\bullet$ ECML still significantly enhance the performance of RMML in all test cases. It verifies the fact that, ECML is an effective ensemble cascade metric learning mechanism both suitable for small-scale and large-scale face data;

$\bullet$ 
RMML outperforms the other scalable metric learning approaches of the closed-form solution. It demonstrates that, RMML is a more suitable metric learning method for face verification when CNN feature is used for face characterization;

$\bullet$ It is worthy noting that, in large-scale test case $M_{y=1}$ and XQDA may be even inferior to the raw CNN feature only with PCA. However, this does not happen to RMML and EC-RMML. In our opinion, the fitting capacity of these metric learning approaches for large-scale face data is not strong enough on CNN feature.

\begin{figure}[t]
	\centering
    \footnotesize
	\begin{minipage}{0.5\linewidth}
		\centerline{\includegraphics[width=4cm]{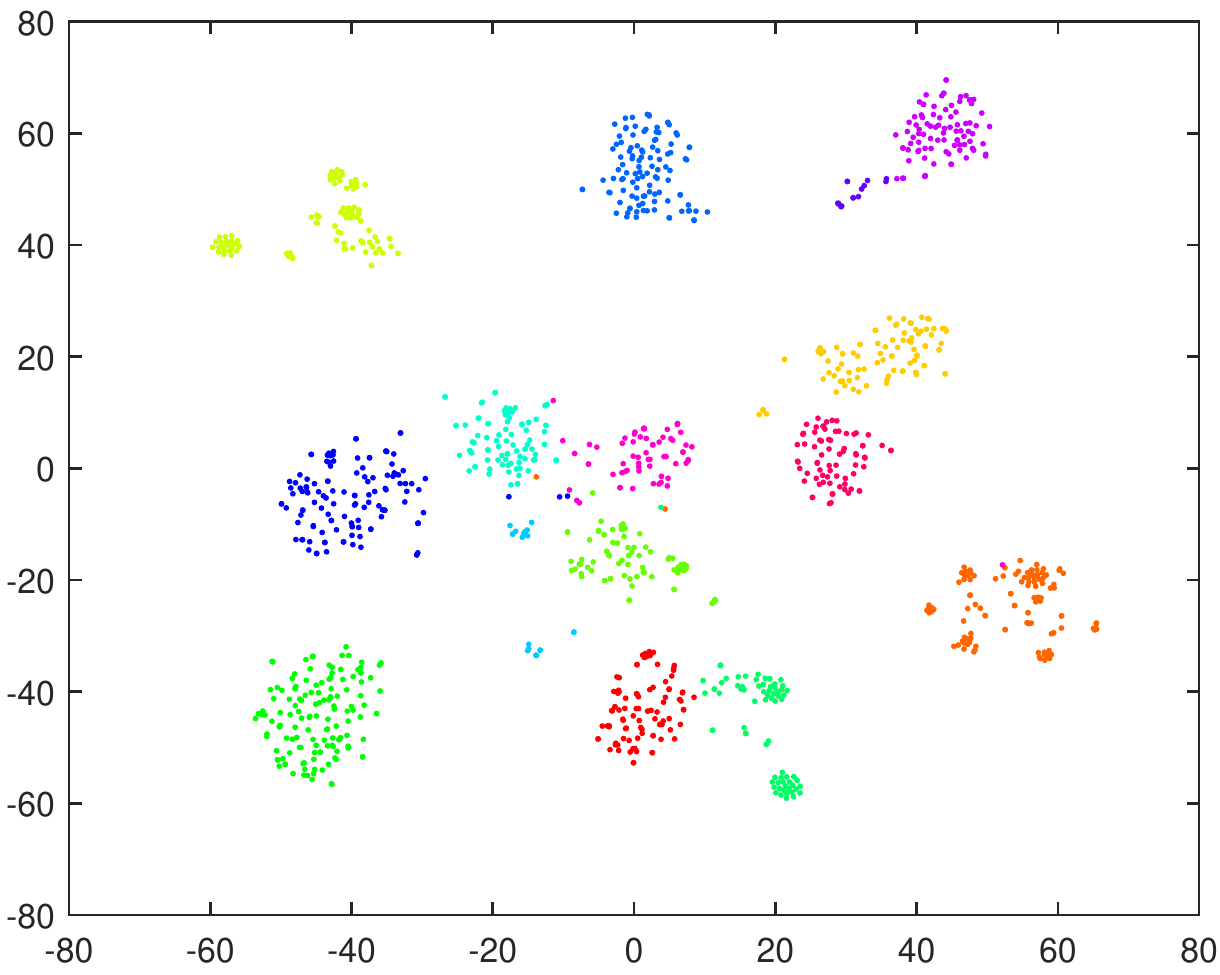}}
		\centerline{(a) Raw CNN feature}
	\end{minipage}
	\begin{minipage}{0.46\linewidth}
		\centerline{\includegraphics[width=4cm]{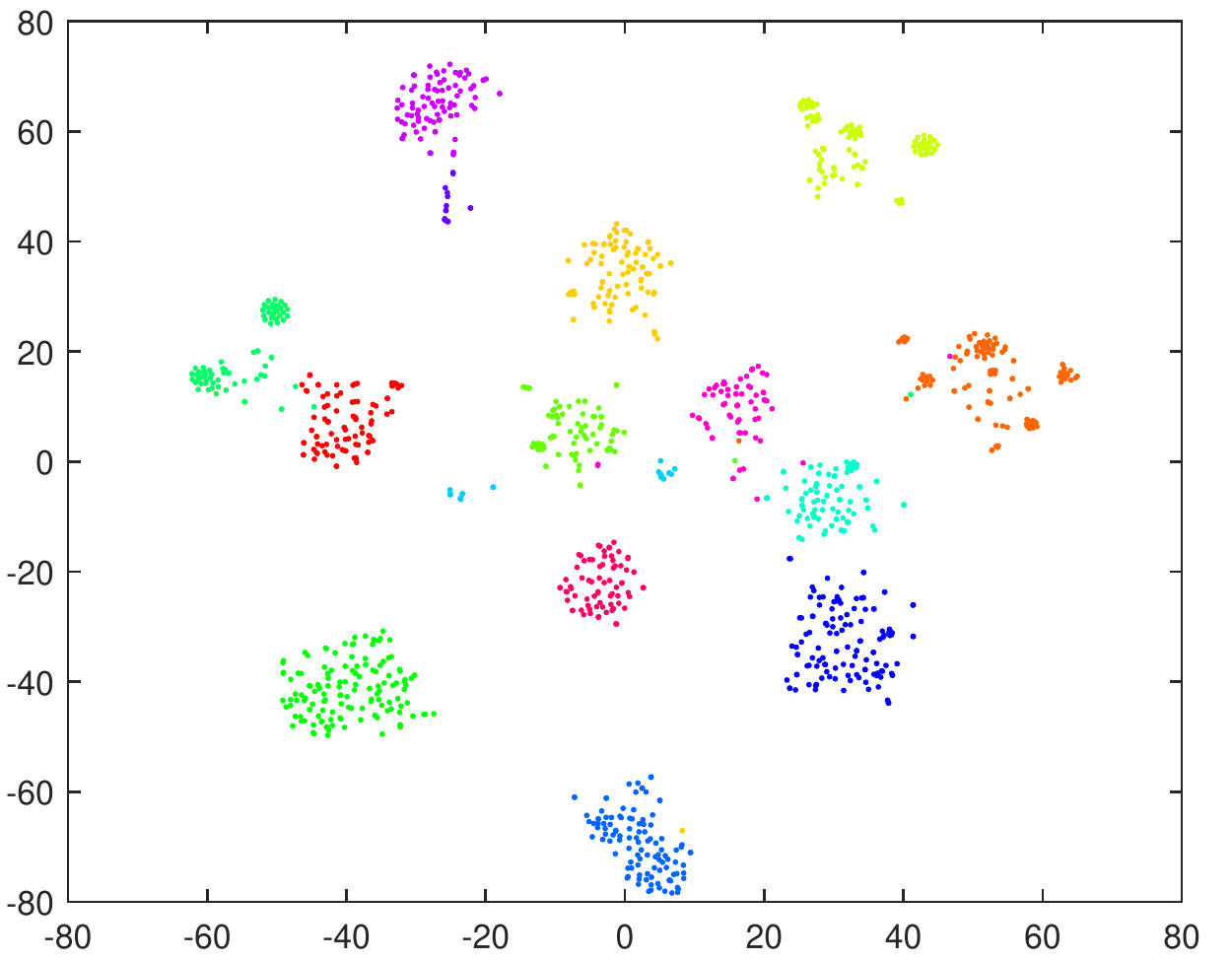}}
		\centerline{(b) With EC-RMML}
	\end{minipage}
	\caption{CNN face feature distribution before and after using EC-RMML. In particular, (a) shows the raw input CNN feature, and (b) shows the output of the second cascade metric learning stage to last within EC-RMML. Each color corresponds to one person identity. It is drawn using t-SNE~{\cite{tsne}}.}
	\label{fig:before_after}
\end{figure}

In addition, to intuitively reveal the effect of EC-RMML towards face verification we draw the CNN face feature distribution before and after using EC-RMML in Fig.~\ref{fig:before_after}. In particular, 1276 face images from 15 persons are involved. Obviously, using EC-RMML the distance between the different persons has been enlarged to essentially improve the discriminative power.

\begin{table}[t]
    \small
    \caption{Performance comparison of EER (\%) among the different metric learning approaches on FV feature, using \textbf{moderatep-scale} test protocol. \textbf{``EC-"}  represents ensemble cascade learning mechanism. The best performance that corresponds to each PCA dimensionality is shown in boldface.}
    \label{table:fv_mediate}
    \centering
    \begin{tabular}{@{\hspace{2mm}}c@{\hspace{2mm}}c@{\hspace{2mm}}c@{\hspace{2mm}}c@{\hspace{2mm}}c@{\hspace{2mm}}c@{\hspace{2mm}}}
    \hline
        PCA dim. &320 &160 &80 &40 \\
        \hline
        Raw feature  &43.24 &43.22  &43.54 &43.58 \\
        $M_{y=1}$~\cite{KISSme} &41.31 &40.51 &41.21 &41.55  \\
        KISSME~\cite{KISSme} &32.21 &31.55 &32.65 &35.72  \\
        SILD~\cite{SILD} &34.00 &33.51 &34.70 &36.61  \\
        XQDA~\cite{XQDA} &28.95 &30.00 &31.54 &34.94  \\
        \hline
        RMML (ours) &36.55 &36.89 &37.73 &38.86  \\
        EC-RMML (ours) &34.31 &35.00 &35.81 &37.45  \\
        \hline
        EC-KISSME (ours) &-- &-- &-- &--\\
        EC-XQDA (ours) &\textbf{24.61} &\textbf{26.29} &\textbf{29.11} &\textbf{33.09}\\
        \hline
        \end{tabular}
        \end{table}

\begin{figure}[t]
\centering
\includegraphics[width=7cm]{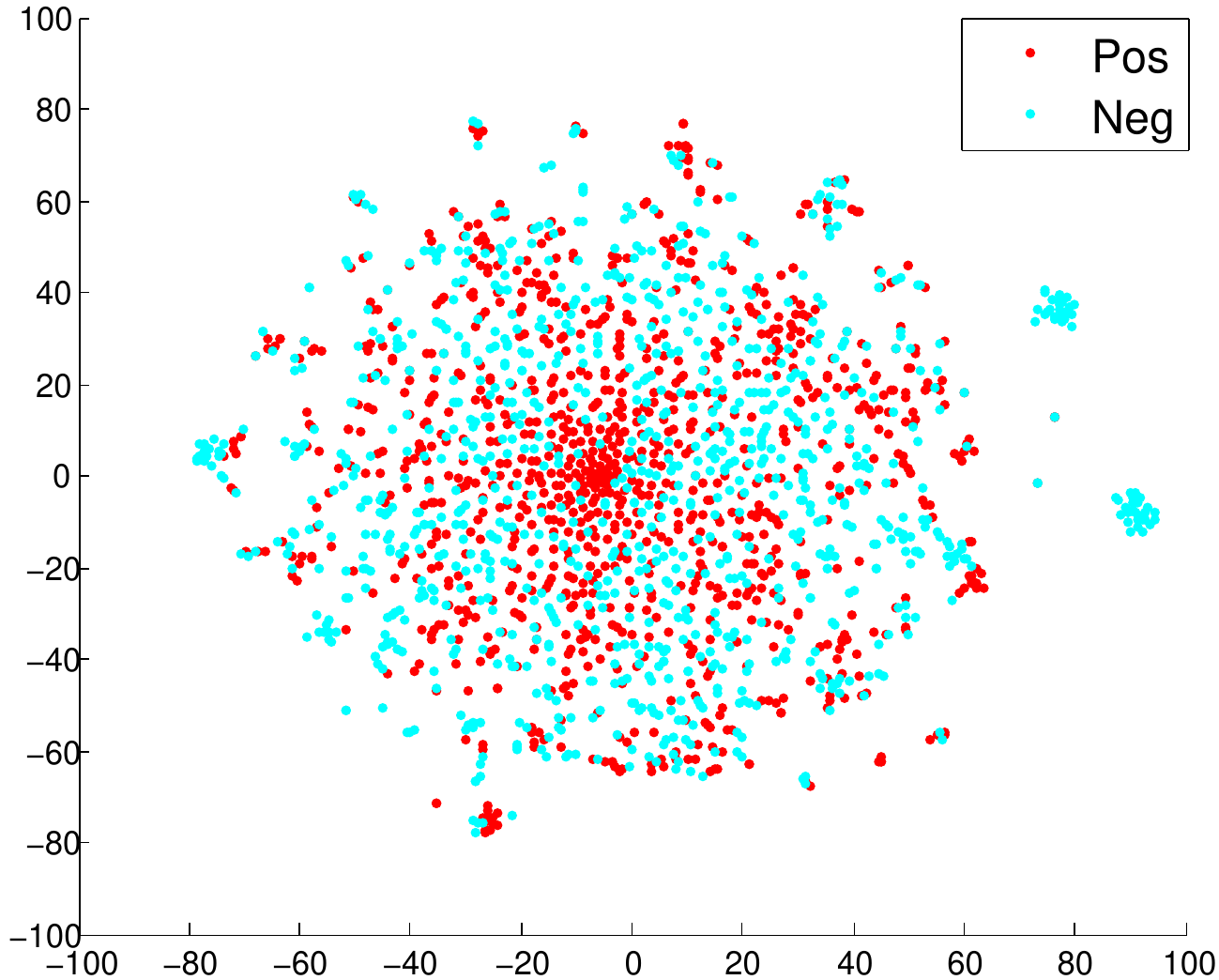}
\caption{Feature difference distribution of the 2,000 matched and unmatched face pairs, when using FV feature. \textbf{``Pos"} denotes the matched case, and \textbf{``Neg"} indicates the unmatched case. It is drawn using t-SNE~\cite{tsne}.}
\label{fig:FVdistribution_2d}
\end{figure}

\subsection{Performance comparison on FV feature} \label{sec:fv_test}

Here, we extract the FV-based face feature following the paradigm in \cite{FVface}. In particular, SIFT is used as the low-level feature and 30 Gaussians are involved in Gaussian mixture model (GMM). Using the moderate-scale test protocol, the performance comparison of EER among the different metric learning approaches is listed in Table~\ref{table:fv_mediate}.  Since the gradient decent based metric learning approaches are extremely time-consuming, we only report the results of the scalable methods of the closed-form or approximate closed-form solution. It can be summarized that:

$\bullet$  Towards FV-based face representation, EC-RMML nor RMML cannot achieve the best performance. They are generally inferior to KISSME and XQDA. The reason seems that, the pairwise FV feature difference distributes in Gaussian form on face, which is beneficial for KISSME and XQDA. To verify this, Fig.~\ref{fig:FVdistribution_2d} shows feature difference distribution of the 2,000 matched and unmatched face pairs using FV feature. Essentially, both of the matched and unmatched cases distributes in Gaussian form approximately. This also justifies our viewpoint in Sec.~\ref{sec:experiment_cnn} that, the performance of metric learning methods relies on the raw feature distribution even for the same visual recognition task;

$\bullet$ ECML still consistently enhances the performance of RMML on FV feature. This demonstrates the effectiveness and generalization capacity of ECML for the different features. Additionally, we apply ECML to KISSME and XQDA. Actually, EC-XQDA outperforms all the other approaches. This reveals that ECML is not only applicable to RMML, which will be further analyzed next;

$\bullet$ Generally speaking, FV feature is inferior to CNN feature for face verification.

\begin{table}[t]
       \small
        \caption{Performance comparison of EER (\%) among the raw KISSME, XQDA, RMML, and their ensemble cascaded versions on CNN feature. \textbf{``EC-"}  represents ensemble cascade learning mechanism.}
        \label{table:ecml_generalization}
        \centering
        \begin{tabular}{@{\hspace{2mm}}c@{\hspace{2mm}}@{\hspace{2mm}}c@{\hspace{2mm}}c@{\hspace{2mm}}c@{\hspace{2mm}}c@{\hspace{2mm}}c@{\hspace{2mm}}c@{\hspace{2mm}}}
        \hline
        ~ & PCA Dim. &640 &320 &160 &80 &40 \\
        \hline
        \multirow{6}{*}{Large-scale}
        &KISSME~\cite{KISSme}  &22.20 &23.07 &21.56 &19.73 &18.45 \\
        &EC-KISSME  &--  &-- &10.14 &10.30 &11.42\\
        \cline{2-7}
         &XQDA~\cite{XQDA} &27.07 &25.98 &23.85 &21.32 &18.97 \\
        &EC-XQDA &31.44  &11.86 &9.83 &10.09 &11.35\\
        \cline{2-7}
         &RMML &18.06 &17.68 &17.41 &17.35 &17.72 \\
        &EC-RMML &12.01  &11.43 &11.20 &11.51 &12.35\\
        \hline
        \multirow{6}{*}{Small-scale}
        &KISSME~\cite{KISSme}  &30.87 &26.79 &25.05 &22.96 &21.74 \\
        &EC-KISSME  &--  &-- &-- &-- &--\\
        \cline{2-7}
         &XQDA~\cite{XQDA} &29.95 &28.37 &25.71 &23.57 &21.79 \\
        &EC-XQDA &33.85  &32.60 &30.99 &14.65 &14.93\\
        \cline{2-7}
        &RMML  &25.44 &23.29 &22.01 &20.76 &20.19 \\
        &EC-RMML  &10.64  &10.47 &10.49 &11.55 &12.66\\
        \hline
        \end{tabular}
\end{table}
\subsection{Discussion on ECML mechanism}

\subsubsection{Generalization capacity of ECML}

In Sec.~\ref{sec:fv_test}, it has been revealed that ECML can also boosted the performance of XQDA on FV feature as well as RMML. Since CNN feature is of stronger face characterization ability, to better justify the generalization capacity of ECML for face verification we summarize the performance of EC-KISSME, EC-XQDA and EC-RMML respectively, using both the large-scale test protocol and the small-scale test protocol. The performance comparison among the raw KISSME and XQDA, and their ensemble cascaded versions is listed in Table~\ref{table:ecml_generalization}. We can see that, if no computation failure occurs, ECML can improve their performance on CNN feature in most cases both for KISSME, XADA and RMML. Actually, the experimental results in Table~\ref{table:CNN_samll},~\ref{table:CNN_large},~\ref{table:fv_mediate} and~\ref{table:ecml_generalization} verify the effectiveness of generalization capacity of ECML to the different metric learning methods and visual features.

\begin{table}[t]
\caption{Average value and standard deviation on EER (\%) of EC-RMML on CNN feature, using \textbf{large-scale} test protocol. In particular, EC-RMML runs for 10 times with the random feature shuffle operation. ``\textbf{Avg.}" indicates the average value of EER, and ``\textbf{Std.}" denotes standard deviation of EER.}
\small
\label{table:Stability}
        \centering
        \begin{tabular}{@{\hspace{2mm}}c@{\hspace{2mm}}c@{\hspace{2mm}}c@{\hspace{2mm}}c@{\hspace{2mm}}c@{\hspace{2mm}}c@{\hspace{2mm}}}
        \hline
        PCA dim. &640 &320 &160 &80 &40\\
        \hline
        RMML &18.06 &17.68 &17.41 &17.35 &17.72\\
        EC-RMML (Avg.)  &12.01  &11.43 &11.20 &11.51 &12.35\\
        EC-RMML (Std.)  &0.52  &0.16 &0.09 &0.08 &0.11\\
        \hline
        \end{tabular}
\end{table}

\subsubsection{Stability of ECML towards random feature shuffle}

Since the random operation that splits the input feature into different groups exists in the ensemble learning procedure of ECML, to verify the stability of ECML we run EC-RMML for 5 times on CNN feature using the large-scale test protocol. Average value and standard deviation on EER of EC-RMML is reported in Table~\ref{table:Stability} to justify the effectiveness and stability of ECML simultaneously. The performance of RMML is also reported. It can be observed that, ECML can significantly enhance the performance of RMML but with low standard deviation (i.e., less than $1\%$). This indeed demonstrates the stability of ECML towards the random feature shuffle operation during the phase of ensemble metric learning.

\begin{table}[t]
       \small
        \caption{Performance comparison of EER (\%) among RMML, its cascaded version and ensemble cascaded version on CNN feature, using \textbf{small-scale} and \textbf{large-scale} test protocol. \textbf{``C-"} indicates cascade learning mechanism, and \textbf{``EC-"}  represents ensemble cascade learning mechanism.}
        \label{table:Ensemble}
        \centering
        \begin{tabular}{@{\hspace{2mm}}c@{\hspace{2mm}}@{\hspace{2mm}}c@{\hspace{2mm}}c@{\hspace{2mm}}c@{\hspace{2mm}}c@{\hspace{2mm}}c@{\hspace{2mm}}c@{\hspace{2mm}}}
        \hline
        ~ & PCA Dim. &640 &320 &160 &80 &40 \\
        \hline
        \multirow{3}*{Large-scale}
        &RMML &18.06 &17.68 &17.41 &17.35 &17.72  \\
        &C-RMML &16.57  &12.56 &11.65 &11.69 &12.27  \\
        &EC-RMML &11.68  &11.43 &11.20 &11.51 &12.35 \\

        \hline
        \multirow{3}*{Small-scale}
        &RMML &25.44 &23.29 &22.01 &20.76 &20.19 \\
        &C-RMML &11.52  &10.65 &10.56 &11.71 &11.58 \\
        &EC-RMML &10.64 &10.47 &10.49 &11.55 &12.66 \\
        \hline
        \end{tabular}
\end{table}

\subsubsection{Effectiveness of cascade metric learning and ensemble metric learning mechanism}

To reveal the effectiveness of cascade metric learning and ensemble metric learning mechanism, we compare the performance of RMML, its cascade boosted version and ensemble cascaded version using CNN feature with the small-scale and large-scale test protocols respectively in Table~\ref{table:Ensemble}. We can see that:

$\bullet$ Both in the large-scale and small-scale test cases, cascade metric learning mechanism consistently improve the performance of RMML by large margins. This essentially justifies the effectiveness of cascade metric learning mechanism;

$\bullet$ Generally, ensemble metric learning mechanism further enhances the performance of the cascade counterpart when the dimensionality is larger than 40. In these cases, the probability of overfitting is relatively higher. This demonstrates that, ensemble metric learning helps to alleviate the overfitting problem that may happen during the phase of cascade metric learning.


\begin{figure}
 \footnotesize
\begin{minipage}{0.48\linewidth}
  \centerline{\includegraphics[width=4.6cm]{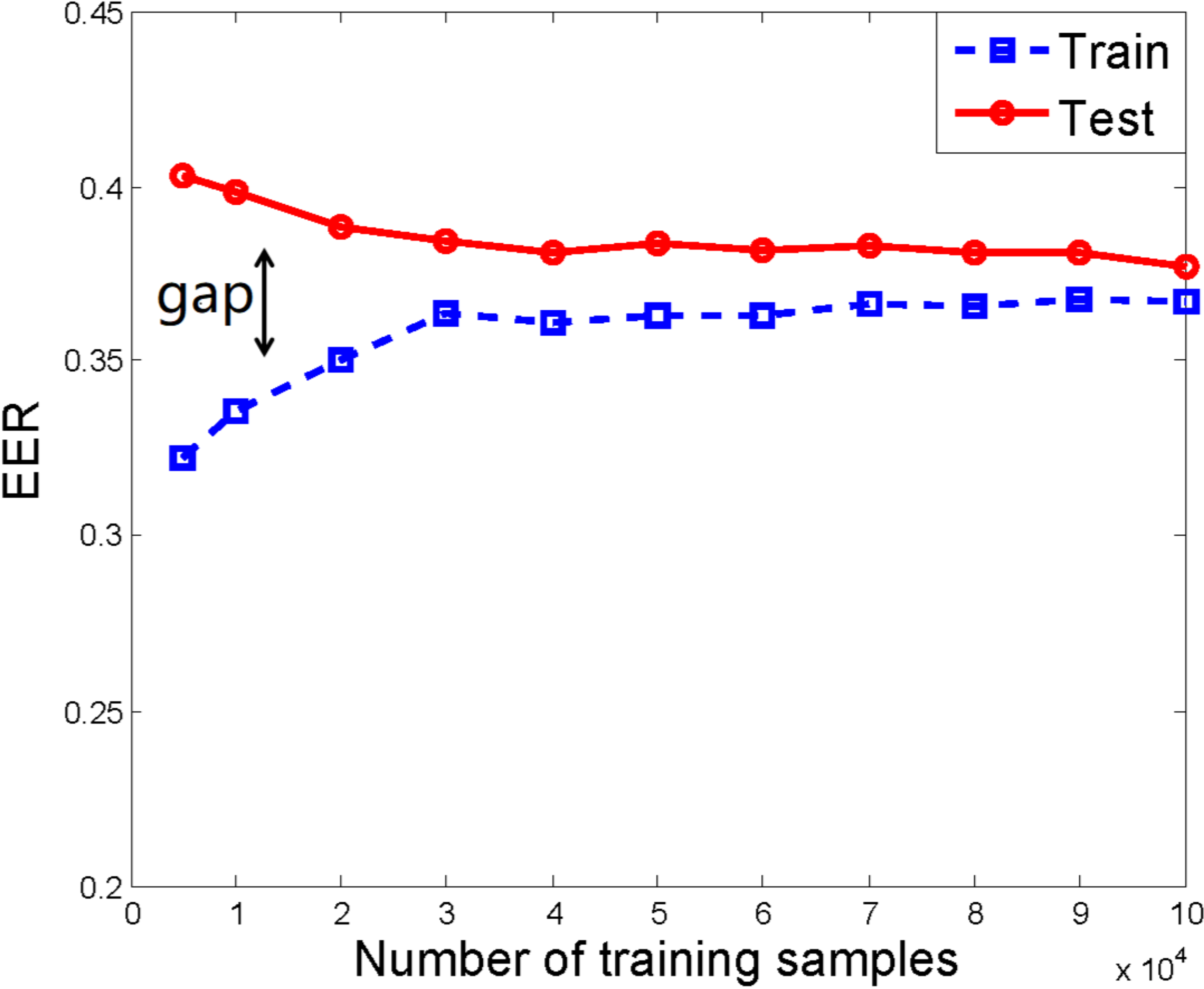}}
  \centerline{(a) RMML}
\end{minipage}
\hfill
\begin{minipage}{.48\linewidth}
  \centerline{\includegraphics[width=4.6cm]{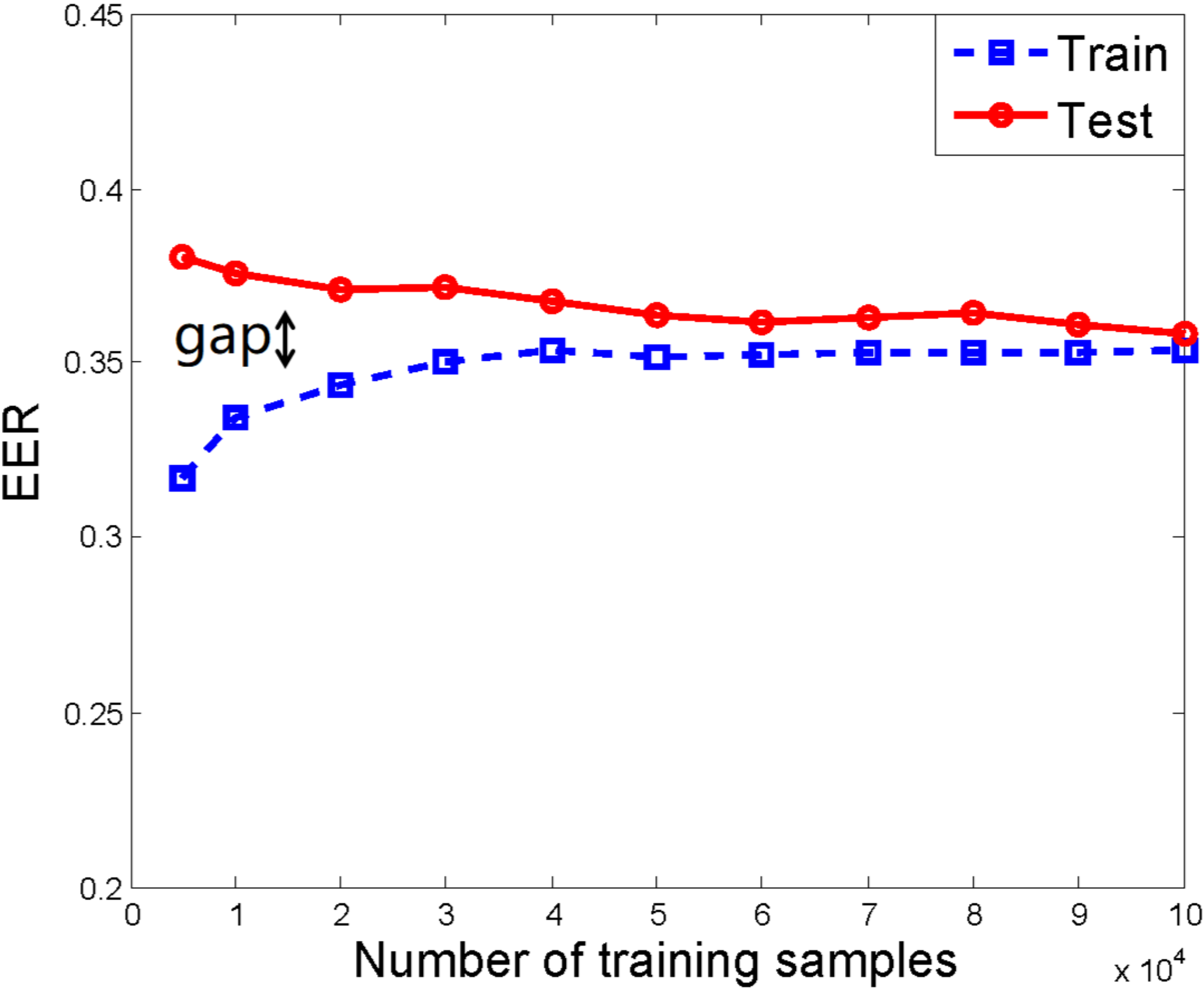}}
  \centerline{(b) EC-RMML}
\end{minipage}
\vfill
\begin{minipage}{0.48\linewidth}

  \centerline{\includegraphics[width=4.6cm]{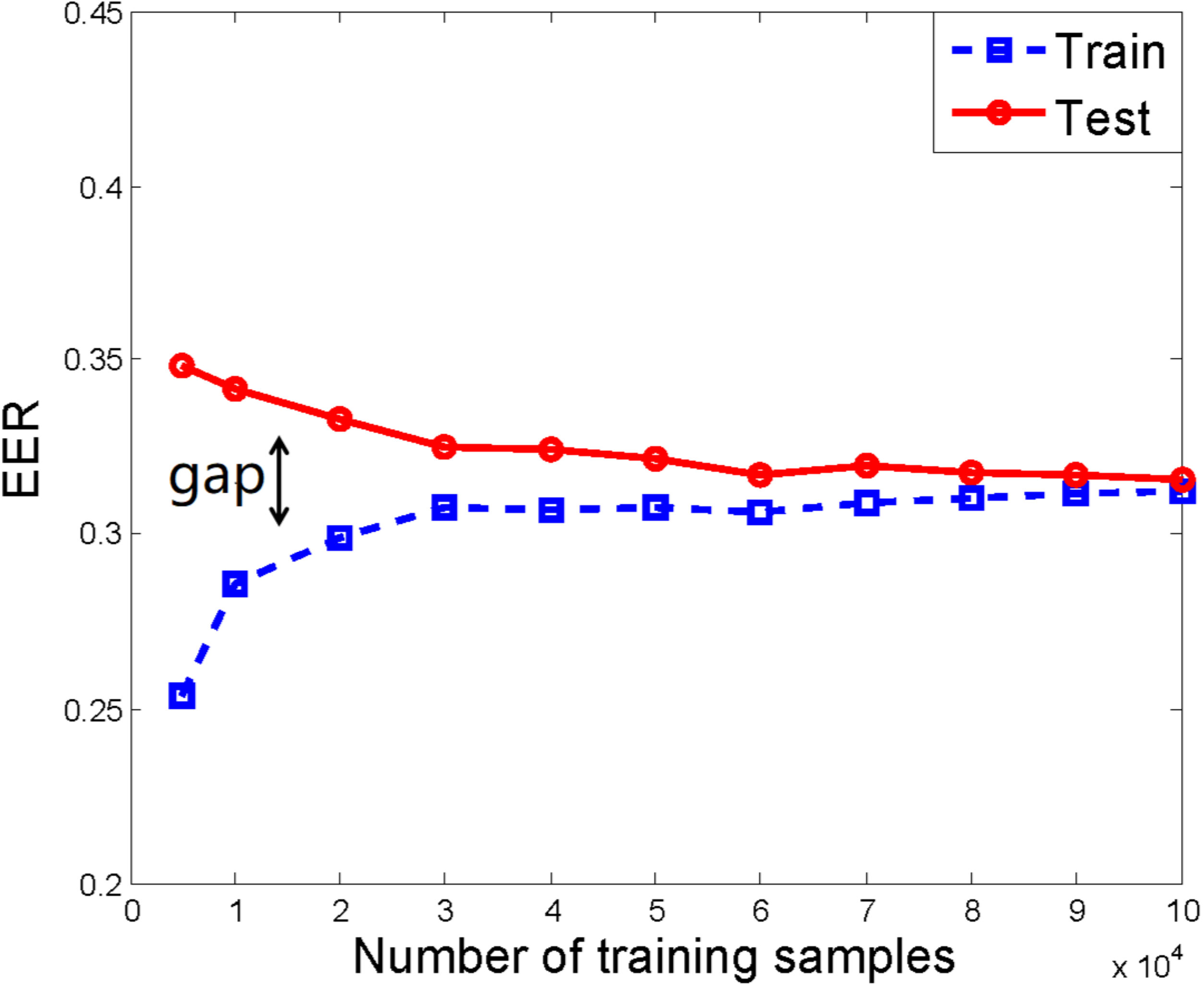}}
  \centerline{(c) XQDA}
\end{minipage}
\hfill
\begin{minipage}{0.48\linewidth}
  \centerline{\includegraphics[width=4.6cm]{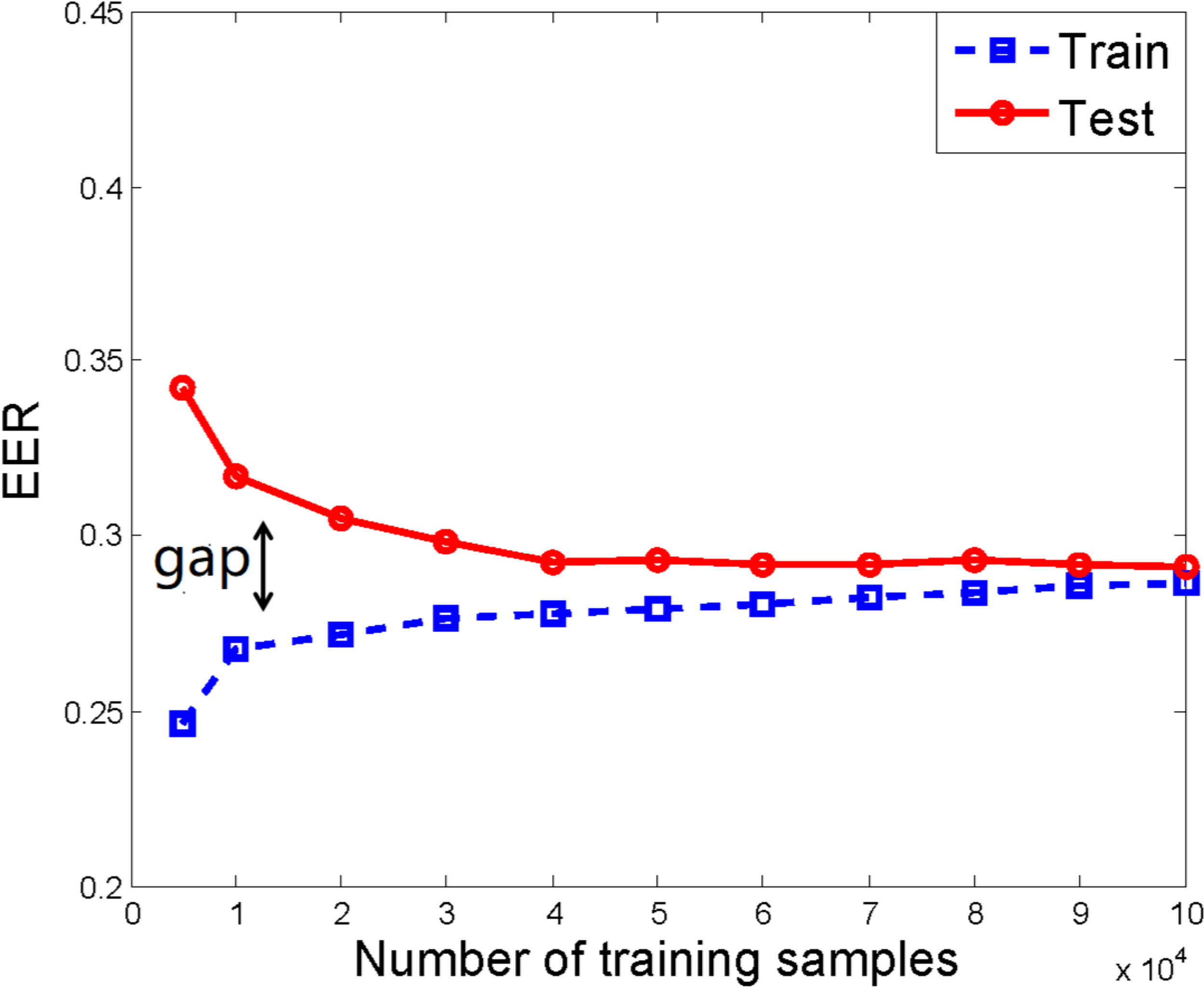}}
  \centerline{(d) EC-XQDA}
\end{minipage}
\caption{The effectiveness of ECML to balance underfitting and overfitting.}
\label{fig:generalization}
\end{figure}
\subsubsection{Good balance between underfitting and overfitting of ECML}

As aforementioned, the key research motivation of ECML is to achieve the good balance between underfitting and overfitting. To reveal this, we conduct the experiments using FV feature of 80 PCA dimensionality with the different amounts of training samples. The test set follows the moderate-scale test protocol setting. The experiments are executed on XQDA and RMML respectively. The training and test EER is reported to simultaneously to reflect the relationship between underfitting and overfitting. The experimental results are shown in Fig.~\ref{fig:generalization}. We can see that:

$\bullet$ XQDA tends to be trapped in overfitting problem. That is, their test EER is much higher than training EER. However when ECML is executed to XQDA, the test EER has been generally reduced. And, the performance gap between training EER and test EER is also lower down. Thus, the discriminative power and generalization capacity of XQDA has been improved by ECML due to its good balance between underfitting and overfitting;

$\bullet$ Using FV feature, RMML generally suffers from underfitting problem. That is, both of the training EER and test EER are high but with the relatively small performance gap. When ECML is applied, both of the training EER and test EER have been reduced. And, the performance gap between training and test EER is not significantly enlarged. This also demonstrates that, ECML actually can maintain the balance between underfitting and overfitting.


\subsubsection{Cascade metric learning stage number setting}

Cascade metric learning mechanism is proposed to improve the fitting capacity of the existing metric learning approaches. Intuitively, when cascade metric learning procedure goes deeper the discriminative power of the yielded distance metric will be further enhanced. But, the overfitting risk will also be increased. Thus, setting the suitable cascade metric learning stage number is an essential issue towards good balance between underfitting and overfitting. To address this, we set the cascade metric learning stage number from 1 to 5, excluding the final metric learning stage. The ensemble cascaded version of RMML (i.e., EC-RMML) runs on CNN feature, using the large-scale and small-scale test protocol respectively. PCA number is set to 640. The performance comparison among the different cascade metric learning stage numbers is listed in Table~\ref{table:cascade_num}. We can see that, with the increment of cascade metric learning stage number the performance of EC-RMML is enhanced. However, when it is too big (e.g., over 3 for large scale) the performance will drop oppositely. That is, the problem of overfitting may happen. Accordingly, the cascade metric learning stage number is empirically set to 3.

\begin{table}[t]
      \small
        \caption{Performance comparison of EER (\%) among the EC-RMMLs with the different cascade metric learning stage numbers on CNN feature, using \textbf{small-scale} and \textbf{large-scale} test protocol. PCA number is set to 640.}
        \centering
        \begin{tabular}{@{\hspace{1.8mm}}c@{\hspace{1.8mm}}@{\hspace{1.8mm}}c@{\hspace{1.8mm}}c@{\hspace{1.8mm}}c@{\hspace{1.8mm}}c@{\hspace{1.8mm}}c@{\hspace{1.8mm}}c@{\hspace{1.8mm}}}
        \hline
        ~ & Cascade No. &1 & 2 &3 &4 &5 \\
        \hline
        \multirow{1}*{Small-scale}
        &EC-RMML &15.85  &12.68 &{10.64} &10.94 &11.34 \\

        \hline
        \multirow{1}*{Large-scale}
        &EC-RMML &13.83  &13.12  &12.01  &12.14  &12.59  \\
        \hline
        \end{tabular}
  \label{table:cascade_num}
\end{table}

\subsection{Discussion on RMML}

\begin{figure}
    \centering
    \footnotesize
	\begin{minipage}{0.5\linewidth}
		\centerline{\includegraphics[width=4.5cm]{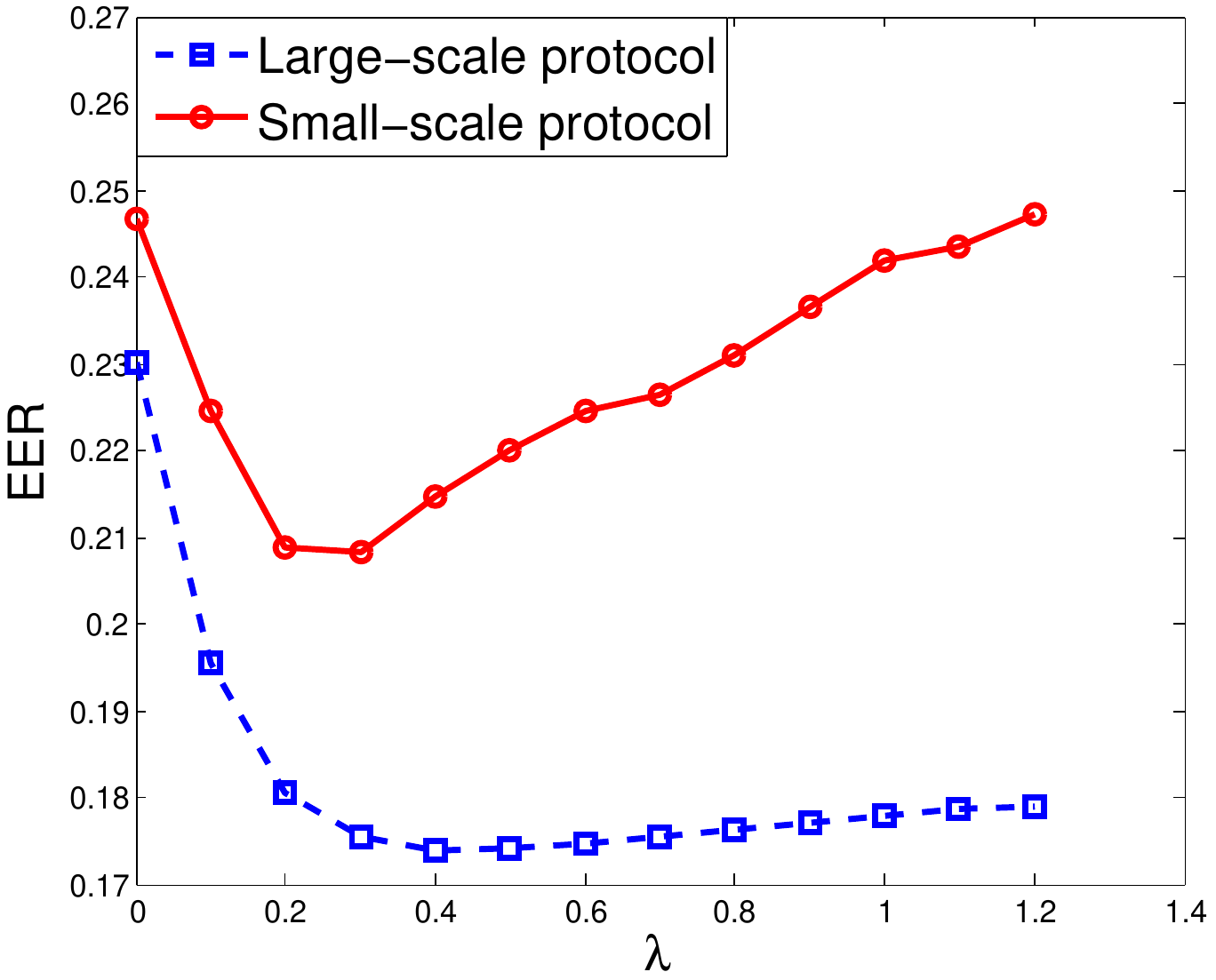}}
		\centerline{(a) RMML}
	\end{minipage}
	\begin{minipage}{0.48\linewidth}
		\centerline{\includegraphics[width=4.5cm]{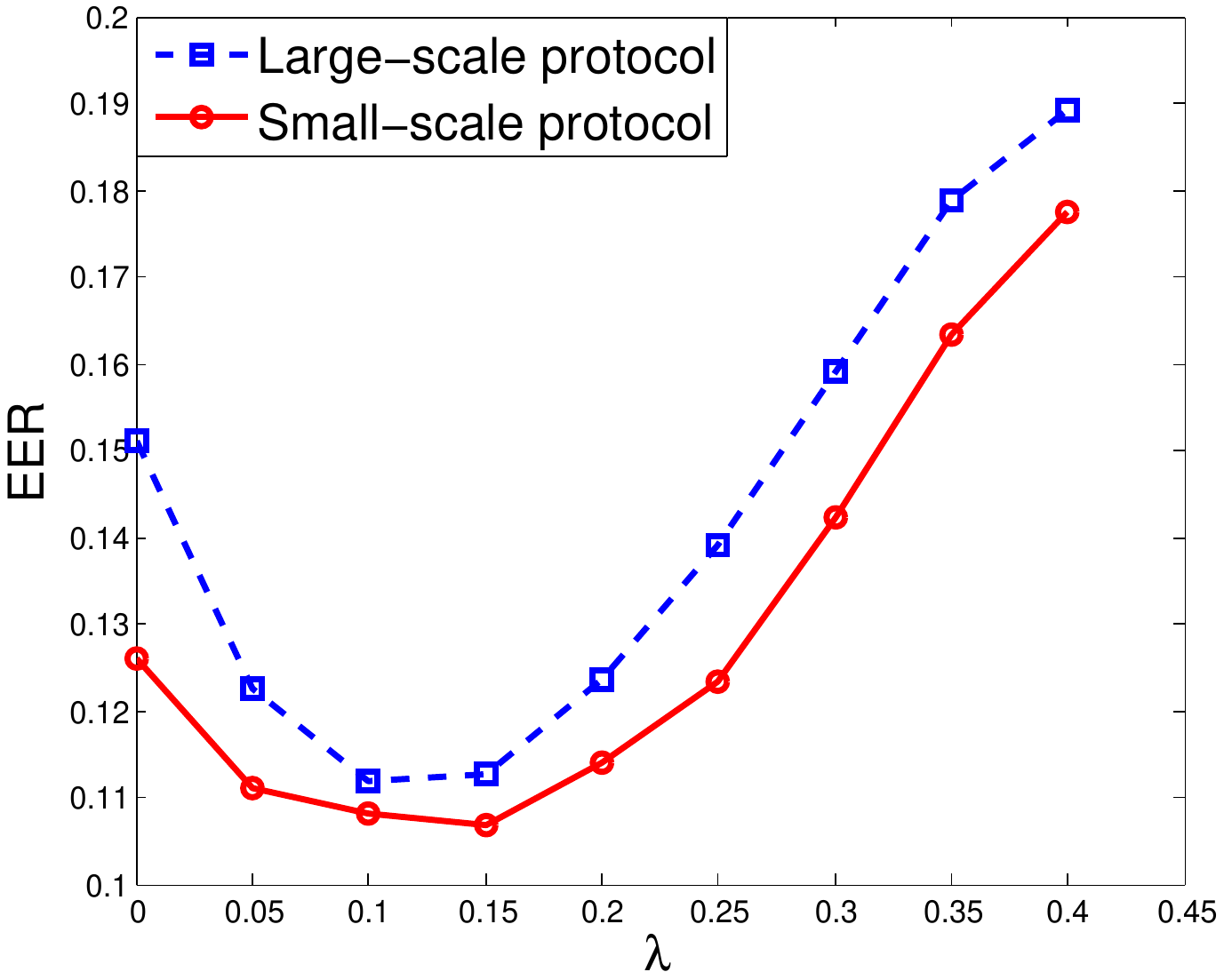}}
		\centerline{(b) EC-RMML}
	\end{minipage}
    \caption{Performance comparison on EER (\%) of RMML and EC-RMML, which corresponds to the different $\lambda$. The experiments are conducted on CNN feature, using the large-scale and small-scale test protocol respectively.}
    \label{fig:lambda}
\end{figure}

\subsubsection{Parameter setting on $\lambda$}

Within RMML, $\lambda$ plays the role of balancing the effect of discriminative term and regularization term. Theoretically, the larger $\lambda$ is the more discriminative the learnt distance metric is, but also suffering from the higher overfitting risk. To choose $\lambda$, we range it from 0 to 1.2 with the stride of 0.1. Fig.~\ref{fig:lambda} (a) shows the performance of RMML with the different $\lambda$ on CNN feature, using the large-scale and small-scale test protocol respectively. PCA number is set to 640. It can be observed that, when $\lambda$ is small (i.e., less than 0.5) with its increment the performance of RMML is enhanced remarkably. However, when $\lambda$ is equal or larger than 0.5 the performance gain is not significant or even with drop. {The reason seems that, when $\lambda$ is too big overfitting problem tends to happen.} Thus, $\lambda$ is set to 0.5 for RMML when it runs solely.

Meanwhile, we also investigate the setting of $\lambda$ for RMML when it is embedded into ECML. Under the same experimental setting of RMML, the performance of EC-RMML that corresponds to the different $\lambda$ is shown in Fig.~\ref{fig:lambda} (b). It can be seen that, with the increment of $\lambda$, the performance of EC-RMML is enhanced. However, when $\lambda$ is too big (e.g., over 0.1 in large-scale protocol) the performance will drop oppositely. Accordingly, $\lambda$ is set to 0.1 for RMML when it is embedded into ECML.

 \subsubsection{Comparison with KISSME}

The proposition of RMML is derived from KISSME. The essential difference between them is that, RMML does not hold the assumption that the pairwise feature difference of the matched and unmatched samples is in Gaussian distribution form. This leads to the fact that, on FV feature that obeys the Gaussian distribution assumption KISSME actually performs better than RMML as shown in Tabel~\ref{table:fv_mediate}. But on CNN feature that does not preserve this distribution property, RMML consistently outperforms RMML as shown in Table~\ref{table:CNN_samll} and~\ref{table:CNN_large}. Since CNN feature is of stronger discriminative power than FV feature for face characterization, we can draw the conclusion that RMML is better choice for face verification than KISSME. Meanwhile, EC-RMML is able to avoid the computation failure problem that may happen to EC-KISSME as shown in Table~\ref{table:CNN_samll}, Table~\ref{table:CNN_large} and Table~\ref{table:fv_mediate}. Thus, RMML is of stronger robustness for the practical applications.

\begin{table}[t]
        \small
        \caption{Running time (s) comparison among the different metric learning approaches. It is conducted on CNN feature, using \textbf{small-scale} test protocol. PCA number is set to 640.}
        \label{table:speed1}
        \centering
        \begin{tabular}{@{\hspace{1.8mm}}c@{\hspace{1.8mm}}c@{\hspace{1.8mm}}c@{\hspace{1.8mm}}c@{\hspace{1.8mm}}c@{\hspace{1.8mm}}}
        \hline
        ~ &KISSME &XQDA &SILD &LMNN\\
        \hline
        Training &0.1710  &0.5988 & 0.6423 & 80.2878 \\
        Test  &0.0021  &0.0020 & 0.0021 &0.0021\\
        \hline
        ~ &ITML &LDML &RMML &EC-RMML\\
        \hline
        Training &2211.4717 & 1234.0512 & 0.1396 & 30.7031\\
        Test &0.0022 & 0.0020 &0.0020 &0.0051\\
        \hline
        \end{tabular}
\end{table}

\subsection{Running time analysis}

{The running time analysis is listed in Tabel~\ref{table:speed1}, Tabel~\ref{table:speed2} and Tabel~\ref{table:speed3} respectively. The run time comparison among the different metric learning methods is listed in Tabel~\ref{table:speed1}. The experiment is conducted on CNN feature using the small-scale test protocol, with the PCA number of 640. In Tabel~\ref{table:speed2} and Tabel~\ref{table:speed3}, we report the training and test time consumption of RMML and EC-RMML over each experimental task. In particular, training time indicates the whole time consumption for training of all samples, and test time denotes the time consumption of testing per sample pair. The raw feature extraction time consumption is excluded. All the experiments run on the computer with Intel (R) Xeon(R) E5-2640 @ 2.00GHz (only using one core) in Matlab. We can see that, due to the closed-form solution both of RMML and EC-RMML essentially run fast, compared to the other metric learning approaches. And, the introduction of ECML towards RMML will not yield heavy extra time consumption. Concerning the remarkable performance gain yielded by ECML, it actually achieves good tradeoff between effectiveness and efficiency.}

\begin{table}[t]
    \newcommand{\tabincell}[2]{\begin{tabular}{@{}#1@{}}#2\end{tabular}}
       \small
        \caption{Training time (s) comparison between RMML and its ensemble cascaded version on FV feature and CNN feature using \textbf{small-scale} and \textbf{large-scale} protocol. \textbf{``EC-"}  represents ensemble cascade learning mechanism.}
        \label{table:speed2}
        \centering
        \begin{tabular}{@{\hspace{0.7mm}}c@{\hspace{0.7mm}}@{\hspace{0.7mm}}c@{\hspace{0.7mm}}c@{\hspace{0.9mm}}c@{\hspace{0.9mm}}c@{\hspace{0.9mm}}c@{\hspace{0.9mm}}c@{\hspace{0.9mm}}}
        \hline
        ~ & PCA Dim. &640 &320 &160 &80 &40 \\
        \hline
        \multirow{2}*{\tabincell{c}{CNN feature\\ Large-scale}}
        &RMML &2.6251 &1.0121 &0.4711 &0.1895 &0.0987  \\
        &EC-RMML &56.3310  &28.6912 &15.4018 &7.9276 &4.1444 \\

        \hline
        \multirow{2}*{\tabincell{c}{CNN feature\\ Small-scale}}
        &RMML &0.1396 &0.0526 &0.0255 &0.0138 &0.0057 \\
        &EC-RMML &30.7031 &17.1108 &8.5276 &4.3750 &2.1462 \\
        \hline
        \multirow{2}*{\tabincell{c}{FV feature}}
        &RMML &-- &0.4101 &0.1712 &0.0876 &0.0381 \\
        &EC-RMML & -- &12.0208 &6.3146 &3.3256 &1.8542  \\
        \hline
        \end{tabular}
\end{table}

\begin{table}[t]
    \newcommand{\tabincell}[2]{\begin{tabular}{@{}#1@{}}#2\end{tabular}}
       \small
        \caption{Test time (s) comparison between RMML and its ensemble cascaded version on FV feature and CNN feature using \textbf{small-scale} and \textbf{large-scale} protocol. \textbf{``EC-"}  represents ensemble cascade learning mechanism.}
        \label{table:speed3}
        \centering
        \begin{tabular}{@{\hspace{0.6mm}}c@{\hspace{0.8mm}}@{\hspace{1.5mm}}c@{\hspace{1.5mm}}c@{\hspace{1.5mm}}c@{\hspace{1.5mm}}c@{\hspace{1.5mm}}c@{\hspace{1.5mm}}c@{\hspace{1.5mm}}}
        \hline
        ~ & PCA Dim. &640 &320 &160 &80 &40 \\
        \hline
        \multirow{2}*{\tabincell{c}{CNN feature\\ Large-scale}}
        &RMML &0.0020 &0.0011 &0.0006 &0.0003 &0.0002 \\
        &EC-RMML &0.0051 &0.0026 &0.0014 &0.0011 &0.0008  \\

        \hline
        \multirow{2}*{\tabincell{c}{CNN feature\\ small-scale}}
        &RMML &0.0020 &0.0010 &0.0005 &0.0003 &0.0002 \\
        &EC-RMML &0.0051 &0.0026 &0.0013 &0.0010 &0.0008 \\
        \hline
        \multirow{2}*{\tabincell{c}{FV\\ feature}}
        &RMML &-- &0.0010 &0.0005 &0.0003 &0.0002\\
        &EC-RMML &-- &0.0026 &0.0013 &0.0010 &0.0008 \\
        \hline
        \end{tabular}
\end{table}

\section{Conclusions}

In this paper, an ensemble cascade metric learning mechanism (ECML) is proposed by us for face verification. Essentially, ECML takes the advantage of achieving good balance between underfitting and overfitting. Specifically, cascade metric learning is executed to boost discriminative power to address underfitting problem. Meanwhile, ensemble metric learning is conducted coordinately to alleviate the underlying overfitting risk. The extensive experiments demonstrate that, ECML can improve the performance of the existing metric learning methods on different visual features, without huge extra computational burden.

A robust Mahalanobis metric learning approach (RMML) of closed-form solution is also proposed. RMML does not require the pairwise feature difference of the matched and unmatched samples distributes in Gaussian form as KISSME. And, it can avoid the potential computation failure problem that happens to KISSME, SILD, XQDA, etc. On CNN feature, RMML and its ensemble cascaded version (EC-RMML) outperform the other metric learning approaches significantly. The running speed is also high.

Currently, the cascade metric learning stage number within ECML is not high due to the overfitting problem. Inspired by the great success of deep residual neural network of hundred layers~\cite{resnet}, in future work we plan to introduce the idea of feature residual to ECML. We wish this helps to deepen the cascade metric learning stage to further enhance the discriminative power of the learnt distance metric.

\section*{Acknowledgment}
This work is jointly supported by National Natural Science Foundation of China (Grant No. 61502187, 61772256, and 61876211), the Equipment Pre-research Field Fund of China (Grant No. 61403120405), National Key R\&D Program of China (No. 2018YFB1004p600), Fundamental Research Funds for the Central Universities (Grant No. 2019kfyXKJC024), and National Key Laboratory Open Fund of China (Grant No. 6142113180211). Joey Tianyi Zhou is supported by Singapore Government's Research, Innovation and Enterprise 2020 Plan (Advanced Manufacturing and Engineering domain) under Grant A18A1b0045.


\small{
\bibliographystyle{IEEEtran}
\bibliography{IEEEabrv,references}}

\begin{IEEEbiography}[{\includegraphics[width=1in,height=1.25in,clip,keepaspectratio]{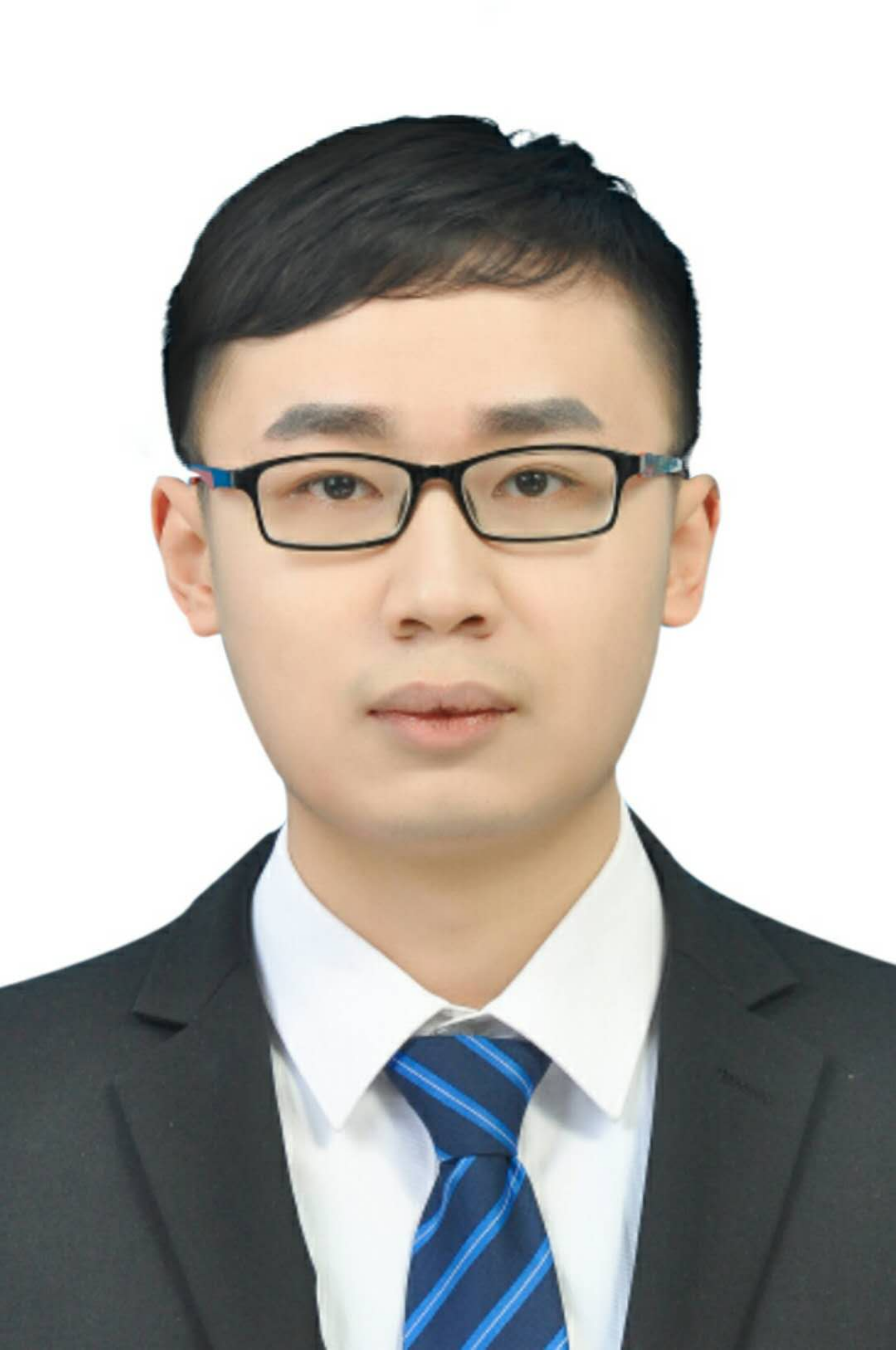}}]{Fu Xiong} recieved the B.S. degree from Huazhong University of Science and Technology, Wuhan, China, in 2016, where he is currently pursuing the M.S. degree with the School of Artificial Intelligence and Automation. His current research interests include human action recognition, face recognition and person re-identification.
\end{IEEEbiography}

\begin{IEEEbiography}[{\includegraphics[width=1in,height=1.25in,clip,keepaspectratio]{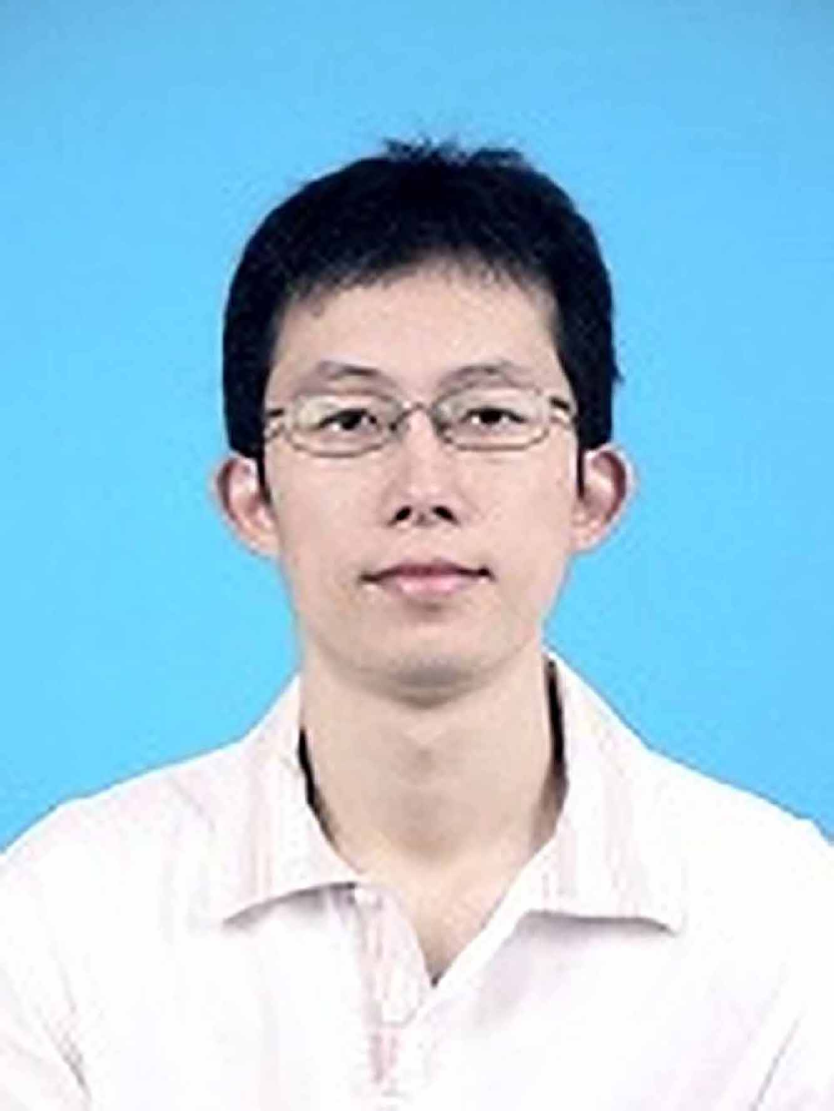}}]{Yang Xiao} received his BS, MS and PhD degrees from Huazhong University of Science and Technology, P.R. China. He is currently an associate professor in the School of Artificial Intelligence and Automation at Huazhong University of Science and Technology, China. Previously, he was ever the research fellow in the School of Computer Engineering and Institute of Media Innovation at Nanyang Technological University, Singapore. His research interests involve computer vision, image processing and machine learning.
\end{IEEEbiography}	

\begin{IEEEbiography}[{\includegraphics[width=1in,height=1.25in,clip,keepaspectratio]{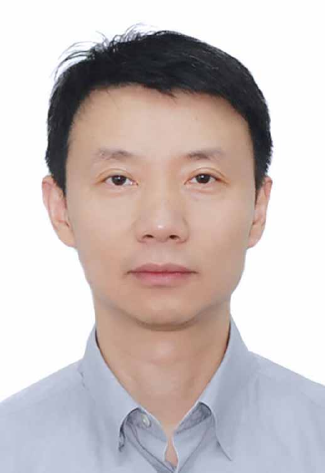}}]{Zhiguo Cao} is a professor of School of Artificial Intelligence and Automation in Huazhong University of Science and Technology. He received his BS and MS degrees in communication and information System from the University of Electronic Science and Technology of China, and his PhD degree in Pattern Recognition and Intelligent System from Huazhong University of Science and Technology. His research interests spread across image understanding and analysis, depth information extraction, 3d video processing, motion detection and human action analysis. His research results, which have published dozens of papers at international journals and prominent conferences, have been applied to automatic observation system for crop growth in agricultural, for weather phenomenon in meteorology and for object recognition in video surveillance system based on computer vision.
\end{IEEEbiography}

\begin{IEEEbiography}[{\includegraphics[width=1in,height=1.25in,clip,keepaspectratio]{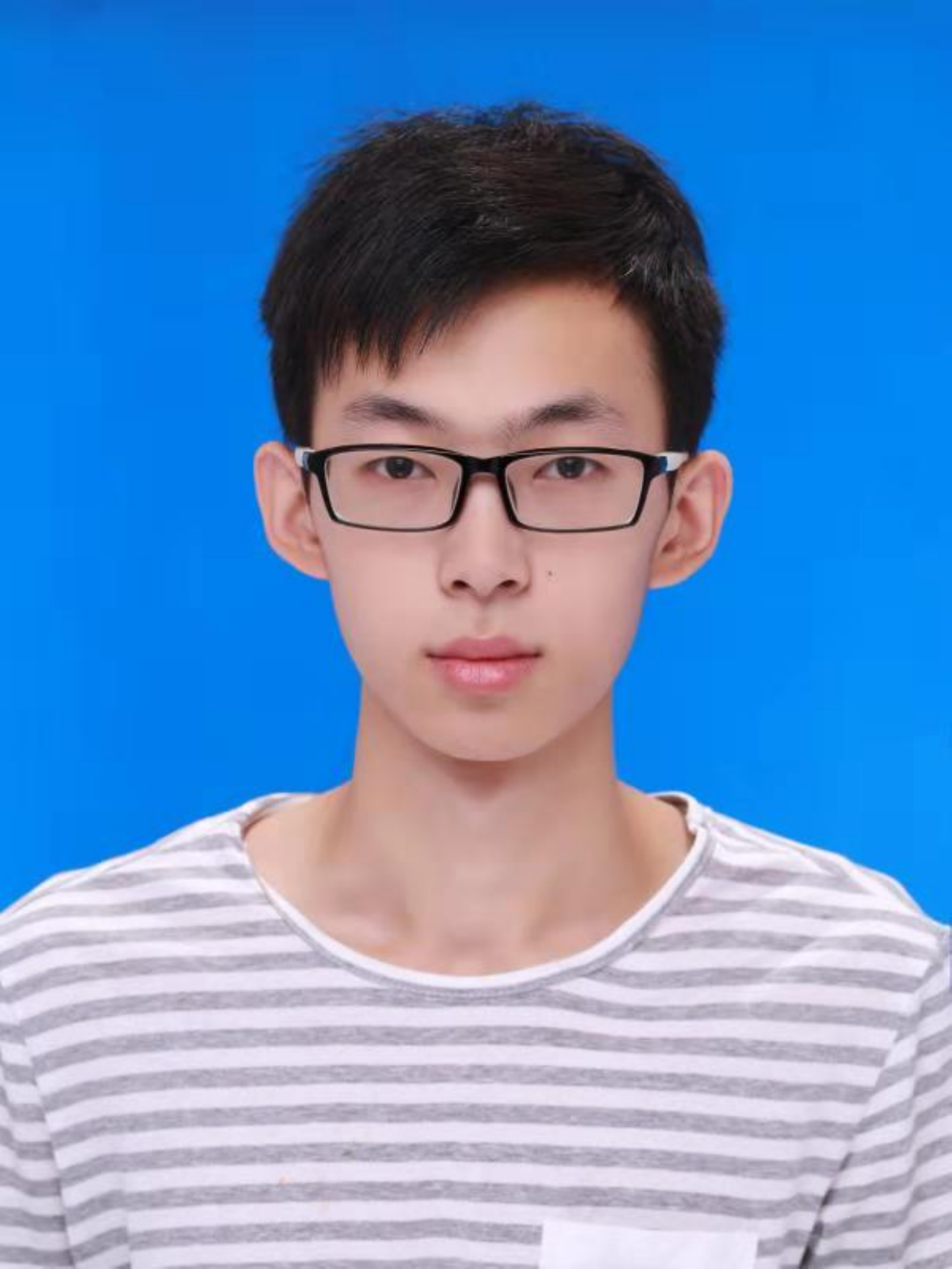}}]{Yancheng Wang} recieved the B.S. degree from Hunan University, China, in 2018. He is currently pursuing the M.S. degree in the School of Artificial Intelligence and Automation of Huazhong University of Science and Technology, Wuhan, China. His current research interests include human action recognition, video understanding and object detection.
\end{IEEEbiography}

\begin{IEEEbiography}[{\includegraphics[width=1in,height=1.25in,clip,keepaspectratio]{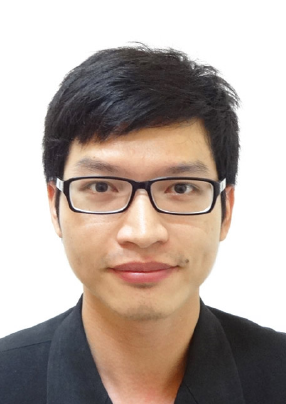}}]{Joey Tianyi Zhou} received the Ph.D. degree in computer science from Nanyang Technological University, Singapore, in 2015.
He is currently a Scientist with the Institute of High Performance Computing, Research Agency for Science, Technology, and Research, Singapore.

Dr. Zhou was a recipient of the Best Poster Honorable Mention at ACML 2012, the Best Paper Award from the BeyondLabeler Workshop on IJCAI 2016, the Best Paper Nomination at ECCV 2016, and the NIPS 2017 Best Reviewer Award. He has served as an Associate Editor for IEEE Access, a Guest Editor for IET Image Processing.
\end{IEEEbiography}

\begin{IEEEbiography}[{\includegraphics[width=1in,height=1.25in,clip,keepaspectratio]{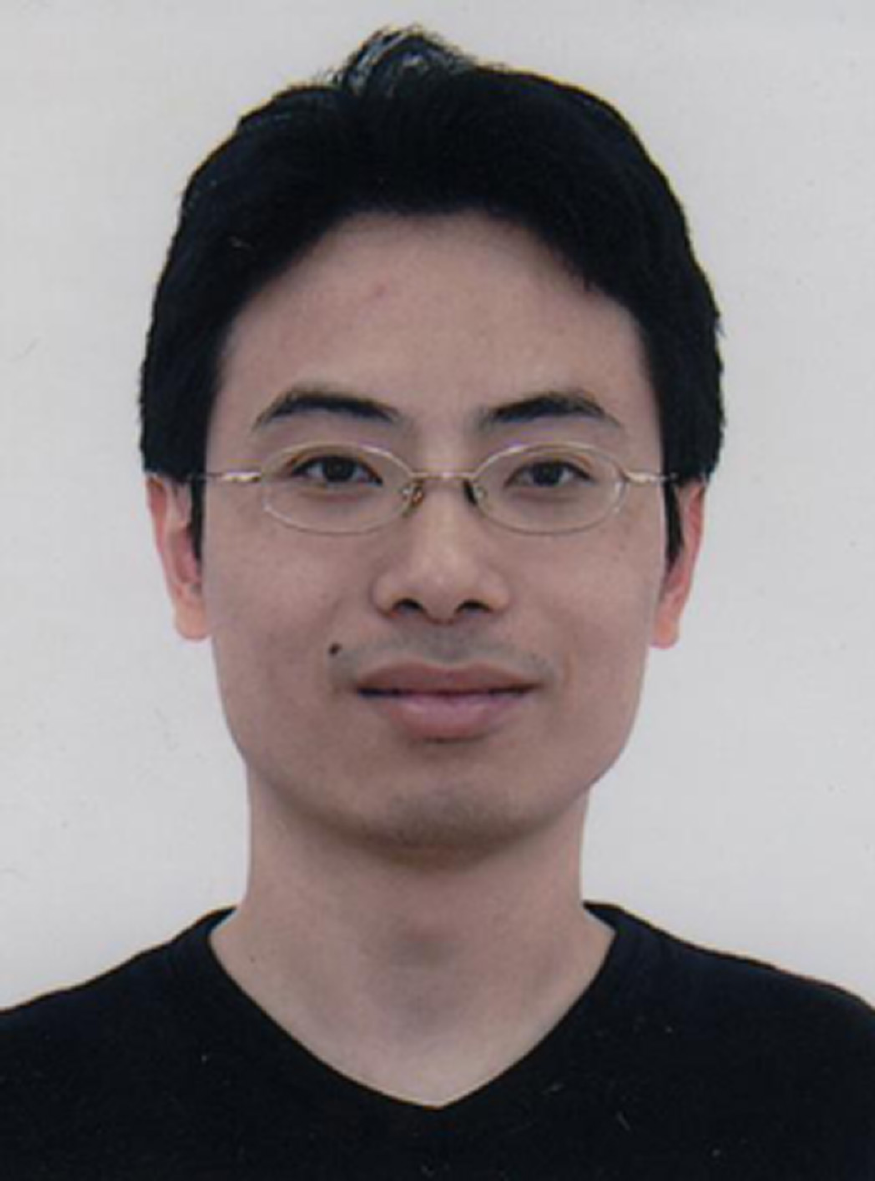}}]{Jianxin Wu} received his BS and MS degrees in computer science from Nanjing University, and his PhD degree in computer science from the Georgia Institute of Technology. He is currently a professor in the Department of Computer Science and Technology at Nanjing University, China, and is associated with the National Key Laboratory for Novel Software Technology, China. He has served as an area chair for CVPR, ICCV, and AAAI, and is an associate editor for the Pattern Recognition Journal. His research interests are computer vision and machine learning. He is a member of the IEEE.
\end{IEEEbiography}

\end{document}